\documentclass[10pt,journal,compsoc]{IEEEtran}

\usepackage{soul}
\usepackage{url}
\usepackage{ulem}
\usepackage{booktabs}
\usepackage{amsmath}
\usepackage{amsthm} 
\usepackage{amsfonts}  
\usepackage{amssymb}
\usepackage{multicol}
\usepackage{multirow}
\usepackage{graphicx}
\usepackage{booktabs}
\usepackage{colortbl}
\usepackage{stfloats}
\usepackage{comment}
\usepackage{makecell}
\usepackage[noend]{algpseudocode}
\usepackage{algorithmicx,algorithm}
\usepackage{pifont}
\usepackage{color}
\usepackage{microtype}
\usepackage{cleveref}
\usepackage[]{footmisc}

\ifCLASSOPTIONcompsoc
  \usepackage[nocompress]{cite}
\else
  \usepackage{cite}
\fi

\begin{document}

\title{Relational Temporal Graph Reasoning for Dual-task Dialogue Language Understanding}

\author{Bowen~Xing and~Ivor~W.~Tsang,~\IEEEmembership{Fellow,~IEEE} 
\IEEEcompsocitemizethanks{\IEEEcompsocthanksitem Bowen Xing is with Australian Artificial Intelligence Institute (AAII), University of Technology Sydney and Centre for Frontier AI Research, A*STAR.\protect\\
E-mail: bwxing714@gmail.com
\IEEEcompsocthanksitem Ivor Tsang is with Centre for Frontier AI Research, A*STAR and AAII, University of Technology. \protect\\
E-mail: ivor\_tsang@ihpc.a-star.edu.sg}
}

\markboth{Journal of \LaTeX\ Class Files,~Vol.~14, No.~8, August~2015}%
{Shell \MakeLowercase{\textit{et al.}}: Bare Demo of IEEEtran.cls for Computer Society Journals}

\IEEEtitleabstractindextext{%
\begin{abstract}
Dual-task dialog language understanding aims to tackle two correlative dialog language understanding tasks simultaneously via leveraging their inherent correlations. 
In this paper, we put forward a new framework, whose core is relational temporal graph reasoning.
We propose a speaker-aware temporal graph (SATG) and a dual-task relational temporal graph (DRTG) to facilitate relational temporal modeling in dialog understanding and dual-task reasoning. 
Besides, different from previous works that only achieve implicit semantics-level interactions, we propose to model the explicit dependencies via integrating \textit{prediction-level interactions}. 
To implement our framework, we first propose a novel model
\textbf{D}ual-t\textbf{A}sk temporal \textbf{R}elational r\textbf{E}current \textbf{R}easoning network (\textbf{DARER}), which first generates the context-, speaker- and temporal-sensitive utterance representations through relational temporal modeling of SATG, then conducts recurrent dual-task relational temporal graph reasoning on DRTG, in which process the estimated label distributions act as key clues in prediction-level interactions.
And the relational temporal modeling in DARER is achieved by relational convolutional networks (RGCNs).
Then we further propose \textbf{Re}lational \textbf{Te}mporal Trans\textbf{former} (\textbf{ReTeFormer}), which achieves fine-grained relational temporal modeling  via Relation- and Structure-aware Disentangled Multi-head Attention.
Accordingly, we propose \textbf{DARER} with \textbf{R}eTeFormer (\textbf{DARER$^\textbf{2}$}), which adopts two variants of ReTeFormer to achieve the relational temporal modeling of SATG and DTRG, respectively.
The extensive experiments on different scenarios verify that our models outperform state-of-the-art models by a large margin. 
Remarkably, on the dialog sentiment classification task in the Mastodon dataset, DARER and DARER$^2$ gain relative improvements of about 28\% and 34\% over the previous best model in terms of F1. 

\end{abstract} 
}

\maketitle

\IEEEdisplaynontitleabstractindextext

\IEEEpeerreviewmaketitle

\IEEEraisesectionheading{\section{Introduction}\label{sec:introduction}}
\IEEEPARstart{D}ialog language understanding \cite{slu} is the fundamental component of the dialogue system. It includes several individual tasks, e.g. dialog sentiment classification, dialog act recognition, slot filling, and (multiple) intent detection.
In recent years, as researchers discover the inherent correlations among some specific task-pair, the joint task which tackles two tasks simultaneously has attracted increasing attention.
For example, dialog sentiment classification (DSC) and dialog act recognition (DAR) are two challenging tasks in dialog systems \cite{empiricalstudyondialog}, while the task of joint DSC and DAR aims to simultaneously predict the sentiment label and act label for each utterance in a dialog \cite{mastodon,dcrnet}.
An example is shown in Table \ref{table: sample}.
To predict the sentiment of $u_b$, besides its \textit{semantics}, its Disagreement act \textit{label} and the Positive sentiment \textit{label} of its \textit{previous} utterance ($u_a$) can provide useful references, which contribute a lot when humans do this task.
This is because the Disagreement act label of $u_b$ denotes it has the opposite opinion with $u_a$, and thus $u_b$ tends to have a Negative sentiment label, the opposite one with $u_a$ (Positive).
Similarly, the opposite sentiment labels of $u_b$ and $u_a$ are helpful to infer the Disagreement act label of $u_b$.
In this paper, we term this process as dual-task reasoning, where there are three key factors: 
1) the semantics of $u_a$ and $u_b$; 2) the temporal relation between $u_a$ and $u_b$; 3) $u_a$'s and $u_b$'s labels for another task.

\begin{table}[t]
\centering
\fontsize{8.5}{11}\selectfont
\caption{A dialog snippet from the Mastodon \cite{mastodon} dataset.}
\setlength{\tabcolsep}{1mm}{
\begin{tabular}{l|l|l}
\toprule
 Utterances& Act& Sentiment \\ \midrule
 \begin{tabular}[c]{@{}l@{}}$u_a$: I highly recommend it. Really awe-\\some progression and added difficulty \end{tabular} 
& Statement & Positive \\\hline
 $u_{b}$: I never have.
&Disagreement &Negative \\
\bottomrule
\end{tabular}}
\label{table: sample}
\end{table}

\begin{figure}[t]
 \centering
 \includegraphics[width = 0.48\textwidth]{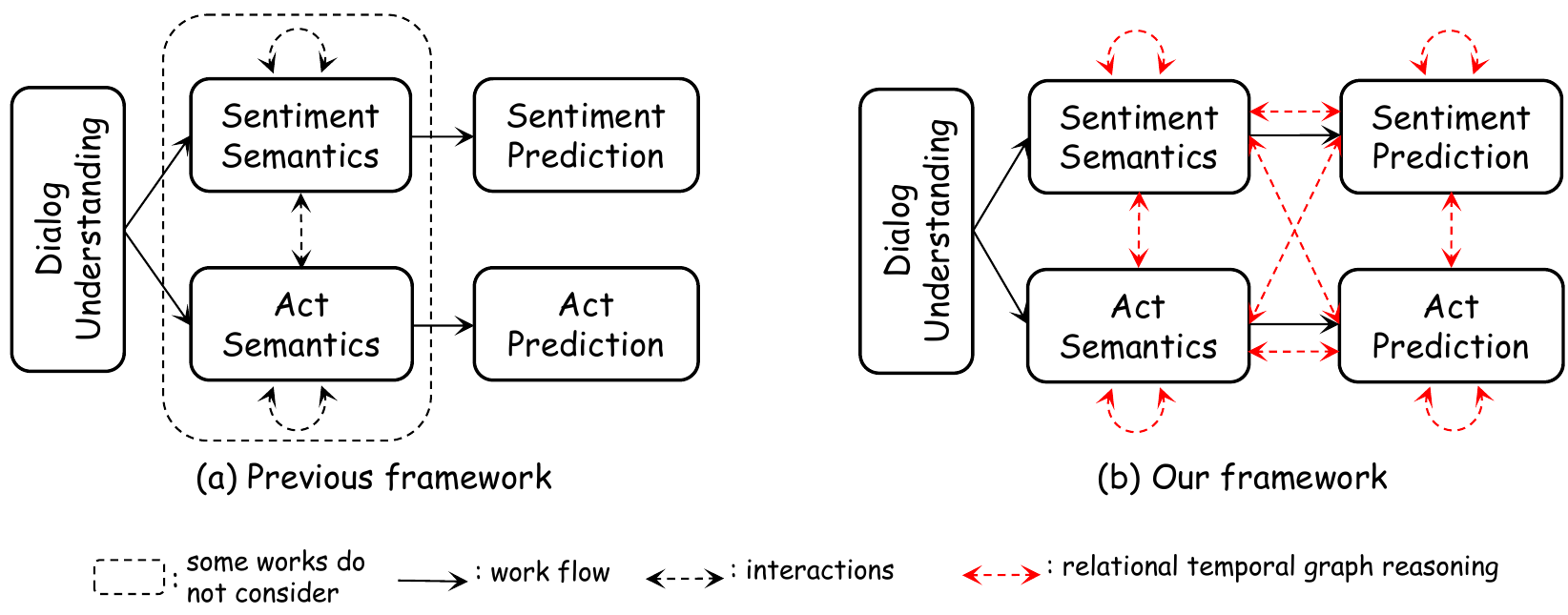}
 \caption{Illustration of previous framework and ours.}
 \label{fig: framework}
\end{figure}

In previous works, different models are proposed to model the correlations between DSC and DAR.
\cite{mastodon} propose a multi-task model in which the two tasks share a single encoder.
\cite{kimkim,dcrnet,bcdcn,cogat} try to model the semantics-level interactions of the two tasks.
The framework of previous models is shown in Fig. \ref{fig: framework} (a).
For dialog understanding, 
Co-GAT \cite{cogat} applies graph attention network (GAT) \cite{gat}
over an undirected disconnected graph which consists of isolated speaker-specific full-connected subgraphs.
Therefore, it suffers from the issue that the inter-speaker interactions cannot be modeled, and the temporal relations between utterances are omitted.
For dual-task reasoning, on the one hand,
previous works only consider the parameter sharing and semantics-level interactions, while the label information is not explicitly integrated into the dual-task interactions.
Consequently, 
the explicit dependencies between the two tasks cannot be captured
and previous dual-task reasoning processes are inconsistent with human intuition, which leverages the label information as crucial clues.
On the other hand, previous works do not consider the temporal relations between utterances in dual-task reasoning, in which they play a key role.

In this paper, we try to address the above issues by introducing temporal relations and leveraging label information. 
To introduce temporal relations, we design a \textbf{s}peaker-\textbf{a}ware \textbf{t}emporal \textbf{g}raph (SATG) for dialog understanding, and a \textbf{d}ual-task \textbf{r}easoning \textbf{t}emporal \textbf{g}raph (DRTG) for dual-task relational reasoning.
Intuitively, different speakers' semantic states will change as the dialog goes, and these semantic state transitions trigger different sentiments and acts.
SATG is designed to model the speaker-aware semantic states transitions, which provide essential indicative semantics for both tasks.
In SATG, there is one group of utterance nodes and two kinds of temporal relations: previous and future.
Since the temporal relation is a key factor in dual-task reasoning,
DRTG is designed to integrate inner- and inter-task temporal relations, making the dual-task reasoning process more rational and effective.
In SATG, there are two parallel groups of utterance nodes and three kinds of temporal relations: previous, future, and equal.


To leverage label information, we propose a new framework, as shown in Fig. \ref{fig: framework} (b).
Except for semantics-level interactions, it integrates several kinds of prediction-level interactions.
First, self-interactions of sentiment predictions and act predictions.
In both tasks, there are prediction-level correlations among the utterances in a dialog.
In the DSC task, the sentiment state of each speaker tends to be stable until the utterances from others trigger the changes \cite{dialoggcn,ercseqtag}.
In the DAR task, there are different patterns (e.g., Questions-Inform and Directives-Commissives) reflecting the interactions between act labels \cite{dailydialog}. 
Second, interactions between the predictions and semantics.
Intuitively, the predictions can offer feedback to semantics, which can rethink and then reversely help revise the predictions.
Third, prediction-prediction interactions between DSC and DAR, which model the explicit dependencies.
However, since our objective is to predict the labels of both tasks, there is no ground-truth label available for prediction-level interactions.
To this end, we design a recurrent dual-task reasoning mechanism that leverages the label distributions estimated in the previous step as prediction clues of the current step for producing new predictions.
In this way, the label distributions of both tasks are gradually improved along the step.
To implement our framework, we propose
\textbf{D}ual-t\textbf{A}sk temporal \textbf{R}elational r\textbf{E}current \textbf{R}easoning Network\footnote{The content of DARER was presented as a poster in ACL 2022 conference.} (DARER) \cite{darer_v1}, which includes three main components.
The \textit{Dialog Understanding} module conducts relation-specific graph transformations (RSGT) over SATG to generate context-, speaker- and temporal-sensitive utterance representations.
The \textit{Initial Estimation} module produces the initial label information which is fed to the \textit{Recurrent Dual-task Reasoning} module, in which RSGT operates on DRTG to conduct dual-task relational reasoning.
And the RSGTs are achieved by relational graph convolutional networks \cite{rgcn}.
Moreover, we design logic-heuristic training objectives to force DSC and DAR to gradually prompt each other in the recurrent dual-task reasoning process.

Then we further propose Relational Temporal Transformer (ReTeFormer) and DARER$^2$. The main difference between DARER and DARER$^2$ is that in DARER$^2$ the original RGCNs applied over SATG and DRTG are replaced with our proposed SAT-ReTeFormer and DTR-ReTeFormer.
The core of ReTeFormer is the Relation- and Structure-Aware Disentangled Multi-head Attention, which can achieve fine-grained relational temporal modeling.
Generally, DARER$^2$ has three distinguished advantages over DARER: (1) ReTeFormer integrates dialog structural information, achieving more comprehensive and fine-grained relational temporal graph reasoning; (2) the relational temporal attention mechanism of ReTeFormer can comprehensively and explicitly model the correlations among dual tasks semantics and predictions; (3) the relation specific attention maps derived by ReTeFormer can provide explainable evidence of relational temporal graph reasoning, making the model more reliable.


In summary, this work has three major contributions. (1) We propose DARER, which is based on a new framework that for the first time achieves relational temporal graph reasoning and prediction-level interactions. (2) Our proposed DARER$^2$ further improves the relational temporal graph reasoning with our proposed ReTeFormer which is based on the Relation- and Structure-Aware Disentangled Multi-head Attention. (3) Experiments prove that DARER and DARER$^2$ significantly outperform the state-of-the-art models in different scenarios of dual-task dialog language understanding.

The remainder of this paper is organized as follows. In Section 2, the related work of two scenarios of dual-task dialog language understanding, (1) Joint Dialog Sentiment Classification and Act Recognition (2) Joint Multiple Intent Detection and Slot Filling, are summarized, and the differences of our method from previous studies are highlighted. Section 3 elaborates on the overall model architecture shared by DARER and DARER$^2$. Section 4 and 5 describe DARER and DARER$^2$, respectively. Experimental results are reported and analyzed in Section 6, and note that the task definition of Joint Multiple Intent Detection and Slot Filling as well as the experiments on this task are introduced in Section 6.10. Finally, we conclude this work and provide some prospective future directions in Section 7.





\section{Related Works}\label{sec: relatedwork}
In recent years, researchers have discovered that some dialog language understanding tasks are correlative and they can be tackled simultaneously by leveraging their beneficial correlations.
\subsection{Joint Dialog Sentiment Classification and Act Recognition}
Dialog Sentiment Classification \cite{cmn,dialoggcn,ket,agru,topicdrivenerc,ercdag} and Dialog Act Recognition \cite{dar_2001,dac_selfatt,speakerawarecrf,Emotionaideddac} are both utterance-level classification tasks.
Recently, it has been found that these two tasks are correlative, and they can work together to indicate the speaker's more comprehensive intentions \cite{kimkim}.
With the development of well-annotated corpora \cite{dailydialog,mastodon}, in which both the act label and sentiment label of each utterance are provided, several models have been proposed to tackle the joint dialog sentiment classification and act recognition task.
Cerisara et al., \cite{mastodon} propose a multi-task framework based on a shared encoder that implicitly models the dual-task correlations.
Kim and Kim \cite{kimkim} integrate the identifications of dialog acts, predictors and sentiments into a unified model.
To explicitly model the mutual interactions between the two tasks,  Qin et al., \cite{dcrnet} propose a stacked co-interactive relation layer, and Li et al., \cite{bcdcn} propose a context-aware dynamic convolution network to capture the crucial local context.
More recently, Qin et al., \cite{cogat} propose Co-GAT, which applies graph attention on a fully-connected undirected graph consisting of two groups of nodes corresponding to the two tasks, respectively.

This work is different from previous works on three aspects.
First, we model the inner- and inter-speaker temporal dependencies for dialog understanding.
Second, we model the cross- and self-task temporal dependencies for dual-task reasoning;
Third, we achieve prediction-level interactions in which the estimated label distributions act as important and explicit clues other than semantics.

\subsection{Joint Multiple Intent Detection and Slot Filling}
It has been widely recognized that intent detection and slot filling have strong correlations.
And a group of models \cite{ijcai2016joint,hakkani2016multi,slot-gated,selfgate,sfid,cmnet,qin2019,jointcap,slotrefine,qin2021icassp,ni2021recent,relanet,coguiding} have been proposed to leverage the correlations for tackling the joint task of intent detection and slot filling in the multi-task framework.
However, in real-word scenarios, a single utterance usually expresses multiple intents, which cannot be handled by the above models.
To this end, Kim et al., \cite{kim2017} propose a multi-intent spoken language understanding model.
Besides, Gangadharaiah and Narayanaswamy \cite{2019-joint} propose the first model that utilizes a slot-gate mechanism to jointly tackle the tasks of multiple intent detection and slot filling.
Furthermore, Qin et al., \cite{agif} propose an adaptive graph-interactive model to model the fine-grained multiple intent information and integrate it into slot filling task via GAT.
More recently, Qin et al., \cite{glgin} propose to conduct non-autoregressive slot decoding in a parallel manner for slot filling, and the proposed GL-GIN achieves state-of-the-art performance.

Our work can be generalized to the joint task of multiple intent detection and slot filling.
Existing methods adopt homogeneous graphs and vanilla GATs to achieve the interactions between the predicted intents and slot semantics, ignoring the specific relations among the two tasks' nodes and the temporal dependencies among the slot nodes.
Different from them, our method achieves relational temporal graph reasoning.


\section{Overall Model Architecture}
\begin{figure*}[t]
 \centering
 \includegraphics[width = \textwidth]{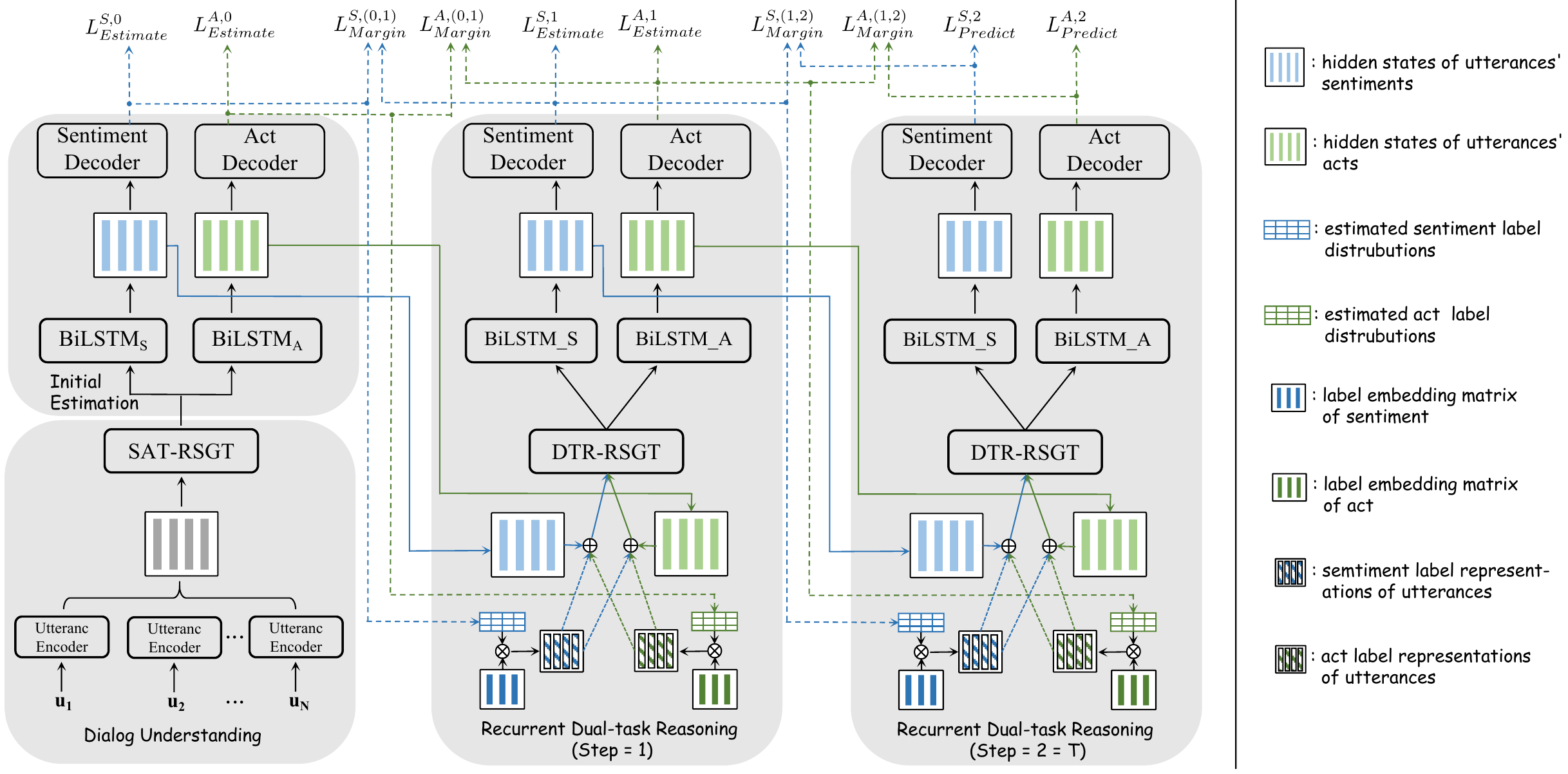}
 \caption{The overall network architecture of DARER and DARER$^2$.
 In DARER, SAT-RSGT and DTR-RSGT are achieved by RGCNs, while in DARER$^2$, they are achieved by SAT-ReTeFormer and DTR-ReTeFormer, respectively. Without loss of generality, the step number $T$ in this illustration is set 2.}
 \label{fig: model}
\end{figure*}
Given a dialog consisting of N utterances: $\mathcal D\!=\!(u_1, u_2, ..., u_N$), our objective is to predict both the dialog sentiment labels $Y^S \!=\! {y^s_1, ..., y^s_N}$ and the dialog act labels $Y^A \!= \!{y^a_1, ..., y^a_N}$ in a single run.

The overall network architecture shared by DARER and DARER$^2$ is shown in Fig. \ref{fig: model}.
It consists of three modules, whose details are introduced in this section.
\subsection{Dialog Understanding}
\subsubsection{Utterance Encoding}
In previous works, BiLSTM \cite{LSTM,aalstm} is widely adopted as the utterance encoder to generate the initial utterance representation: $H=(h_0,...,h_N)$.
In this paper, besides BiLSTM, we also study the effect of different pre-trained language model (PTLM) encoders in Sec. \ref{sec: ptlm}.\\
\textbf{BiLSTM:} We apply the BiLSTM over the word embeddings of $u_t$ to capture the inner-sentence dependencies and temporal relationships among the words, producing a series of hidden states $H_{u,i} = (h_{u,i}^0, ..., h_{u,i}^{l_i})$, where $l_i$ is the length of $u_i$.
Then we feed $H_{u,i}$ into a max-pooling layer to get the representation for each $u_i$.\\
\textbf{PTLM:} We separately feed each utterance into the PTLM encoder and take the output hidden state of the \texttt{[CLS]} token as the utterance representation.

\subsubsection{Speaker-aware Temporal RSGT} \label{sec: sat-rsgt}
To capture the inter- and intra-speaker semantic interactions and the speaker-aware temporal dependencies between utterances, we conduct Speaker-aware Temporal relation-specific graph transformations (SAT-RSGT). 
Now we obtain the context-, speaker- and temporal-sensitive utterance representations: $\hat{H}=(\hat{h_0},...,\hat{h_N})$.
\subsection{Initial Estimation}
To obtain task-specific utterances representations, we separately apply two BiLSTMs over $\hat{H}$ to obtain the utterance hidden states for sentiments and acts respectively: $H_s^0=\text{BiLSTM}_\text{S}(\hat{H})$, $H_a^0=\text{BiLSTM}_\text{A}(\hat{H})$, where $H_s^0=\{h_{s,i}^0\}_{i=1}^{N}$ and $H_a^0=\{h_{a,i}^0\}_{i=1}^{N}.$
Then $H_s^0$ and $H_a^0$ are separately fed into Sentiment Decoder and Act Decoder to produce the initial estimated label distributions:
\begin{equation}
 \begin{aligned}
  P^0_{S}&=\{ P^0_{S,i}\}_{i=1}^N \ P^0_{A}=\{ P^0_{A,i}\}_{i=1}^N\\
 P^0_{S,i} &= softmax(W_d^s h^0_{a,i} + b_d^s)\\
 &= \left[p^0_{s,i}[0], ..., p^0_{s,i}[k], ..., p^0_{s,i}(\left|\mathcal{C}_s\right|-1)\right]\\
 P^0_{A,i} 
 &= softmax(W_d^a h^0_{s,i} + b^a_d)\\
 &=  \left[p^0_{a,i}[0], ..., p^0_{a,i}[k], ..., p^0_{a,i}(\left|\mathcal{C}_a\right|-1)\right]
\end{aligned}
\end{equation}
where $W_d^*$ and $b_d^*$ are weight matrices and biases, $\mathcal{C}_s$ and $\mathcal{C}_a$ are sentiment class set and act class set.

\subsection{Recurrent Dual-task Reasoning}
At step $t$, the recurrent dual-task reasoning module takes two streams of inputs: 1) hidden states $H_s^{t-1}\in\mathbb{R}^{N\times d}$ and $H_a^{t-1}\in\mathbb{R}^{N\times d}$; 2) label distributions $P^{t-1}_{S}\in\mathbb{R}^{N \times \left|\mathcal{C}_s\right|}$ and $P^{t-1}_{A}\in\mathbb{R}^{N \times \left|\mathcal{C}_a\right|}$.

\subsubsection{Projection of Label Distribution}
To achieve the prediction-level interactions, we should represent the label information in vector form to let it participate in calculations.
We use $P_S^{t-1}$ and $P_A^{t-1}$ to respectively multiply the sentiment label embedding matrix $M^e_s\in\!\mathbb{R}^{\left|\mathcal{C}_s\right|\times d}$ and the act label embedding matrix $M^e_a\in\! \mathbb{R}^{\left|\mathcal{C}_a\right|\times d}$, obtaining the sentiment label representations $E_S^{t}=\{e_{s,i}^t\}_{i=1}^N$ and act label representations $E_A^{t}=\{e_{a,i}^t\}_{i=1}^N$.
In particular, for each utterance, its sentiment label representation and act label representation are computed as:
\begin{equation}
 \begin{aligned}
e_{s,i}^t =& \sum_{k=0}^{\left|\mathcal{C}_s\right|-1} p^{t-1}_{s,i}[k] \cdot v_s^k\\
 e_{a,i}^t =& \sum_{k'=0}^{\left|\mathcal{C}_a\right|-1} p^{t-1}_{a,i}[k'] \cdot v_a^{k'}
\end{aligned}
\end{equation}
where $v_s^k$ and $v_a^{k'}$ are the label embeddings of sentiment class $k$ and act class $k'$, respectively.

\subsubsection{Dual-task Reasoning RSGT} \label{sec: dtr-rsgt}
To achieve the self- and mutual-interactions between the semantics and predictions, for each node in DRTG, we superimpose its corresponding utterance's 
label representations of both tasks on its hidden state: 
\begin{equation}
 \begin{aligned}
\hat{h}_{s,i}^t =& h_{s,i}^{t-1} + e_{s,i}^t + e_{a,i}^t \\ 
\hat{h}_{a,i}^t =& h_{a,i}^{t-1} + e_{s,i}^t + e_{a,i}^t 
\end{aligned}
\end{equation}
Thus the representation of each node contains the task-specific semantic features and both tasks' label information, which are then incorporated into the relational reasoning process to achieve semantics-level and prediction-level interactions. 

The obtained $\mathbf{\hat{H}_s^t}$ 
and $\mathbf{\hat{H}_a^t}$ 
both have $N$ vectors, respectively corresponding to the $N$ sentiment nodes and $N$ act nodes on DRTG.
Then we feed them into the Dual-task Reasoning relation-specific graph transformations (DTR-RSGT) conducted on DRTG.
Now we get $\overline{H}_s^t$ and $\overline{H}_a^t$.

\subsubsection{Label Decoding}
For each task, we use a task-specific BiLSTM (TS-LSTM) to generate a new series of task-specific hidden states:
\begin{equation}
 \begin{aligned}
{H}_s^t&=\text{BiLSTM}_\text{S}(\overline{H}_s^t) \\
{H}_a^t&= \text{BiLSTM}_\text{A}(\overline{H}_a^t)
\end{aligned}
\end{equation}
Besides, as $\overline{H}_s^t$ and $\overline{H}_a^t$ both contain the label information of the two tasks, the two TS-$\text{LSTM}$s have another advantage of label-aware sequence reasoning, which has been proven can be achieved by LSTM \cite{lstmsoftmax}. 

Then ${H}_S^t$ and ${H}_A^t$ are separately fed to Sentiment Decoder and Act Decoder to produce $P^{t}_{S}$ and $P^{t}_{A}$.

\subsection{Training Objective}
Intuitively, there are two important logic rules in our model.
First, the produced label distributions should be good enough to provide useful label information for the next step.
Otherwise, noisy label information would be introduced, misleading the dual-task reasoning.
Second, both tasks are supposed to learn more and more beneficial knowledge from each other in the recurrent dual-task reasoning process. Scilicet the estimated label distributions should be gradually improved along steps.
In order to force our model to obey these two rules, we propose a constraint loss $L_{Constraint}$ that includes two terms: $L_{Estimate}$ and $L_{Margin}$,  which correspond to the two rules, respectively.

\textbf{Estimate Loss}
$L_{Estimate}$ is the cross-entropy loss forcing model to provide good enough label distributions for the next step.
At step $t$, for DSC task, $\mathcal{L}^{S,t}_{Estimate}$ is defined as:
\begin{equation}
\mathcal{L}^{S,t}_{Estimate}=\sum_{i=1}^{N}\sum_{k=0}^{\left|\mathcal{C}_s\right|-1}y_i^s[k] \text{log}\left(p_{s,i}^{t}[k]\right) \label{eq: estloss}
\end{equation}

\textbf{Margin Loss}
$L_{Margin}$ works on the label distributions of two adjacent steps, and it promotes the two tasks gradually learning beneficial knowledge from each other via forcing DARER to produce better predictions at step $t$ than step $t-1$.
Besides, although the model can receive more information at step $t$ than $t-1$, this information is imperfect because there are some incorrect predictions of the previous step. Therefore, we use the margin loss to force the model to leverage the \textit{beneficial} information to output better predictions.
For DSC task, $\mathcal{L}^{S,(t,t-1)}_{Margin}$ is a margin loss defined as:
\begin{equation}
\mathcal{L}^{S,(t,t-1)}_{Margin}\!=\!\sum_{i=1}^{N}\sum_{k=0}^{\left|\mathcal{C}_s\right|-1}\!\!y_i^s[k] \ \text{max}(0,p_{s,i}^{t-1}[k]-p_{s,i}^{t}[k])\label{eq: marginloss}
\end{equation}
If the correct class’s possibility at step t is worse than at step t-1, $mathcal{L}^{S,(t,t-1)}_{Margin} > 0$. Then the negative gradient further force the model to predict better at step t. Otherwise, the correct class’s possibility at step t is better than or equal to the one at step t-1. In this case,  $mathcal{L}^{S,(t,t-1)}_{Margin} = 0$.

\textbf{Constraint loss}
$L_{Constraint}$ is the weighted sum of $L_{Estimate}$ and  $L_{Margin}$, with a hyper-parameter $\gamma$ balancing the two kinds of punishments.
For DSC task, $\mathcal{L}^S_{Constraint}$ is defined as:
\begin{equation}
\mathcal{L}^S_{Constraint}=\sum_{t=0}^{T-1}\mathcal{L}^{S,t}_{Estimate} + \gamma_s * \sum_{t=1}^{T}\mathcal{L}^{S,(t,t-1)}_{Margin}\label{eq: Constraintloss}
\end{equation}

\textbf{Final Training Objective}
The total loss for DSC task ($\mathcal{L}^S$) is the sum of $\mathcal{L}^S_{Constraint}$ and $\mathcal{L}^S_{Prediction}$:
\begin{equation}
 \mathcal{L}^S =\mathcal{L}^S_{Prediction} + \mathcal{L}^S_{Constraint} \label{eq: sentiloss}
\end{equation}
where $\mathcal{L}^S_{Prediction}$ is the cross-entropy loss of the produced label distributions at the final step $T$:
\begin{equation}
\mathcal{L}^S_{Prediction}=\sum_{i=1}^{N}\sum_{k=0}^{\left|\mathcal{C}_s\right|-1}y_{s,i}\ \text{log}\left(p_{s,i}^{T}[k]\right) \label{eq: predloss}
\end{equation}

The total loss of DAR task ($\mathcal{L}^A$) can be derivated similarly like \cref{eq: estloss,eq: marginloss,eq: Constraintloss,eq: sentiloss,eq: predloss}.

The final training objective of our model is the sum of the total losses of the two tasks:
\begin{equation}
\mathcal{L} = \mathcal{L}^S+ \mathcal{L}^A
\end{equation}

\section{DARER}
Based on the overall model introduced in Section 3, DARER achieves SAT-RSGT and DTR-RSGT via applying RGCNs on the speaker-aware temporal graph (SATG) and dual-task reasoning temporal graph (DRTG), respectively.
\subsection{SAT-RSGT}
\subsubsection{Speaker-aware Temporal Graph}
\begin{figure}[t]
 \centering
 \includegraphics[width = 0.3\textwidth]{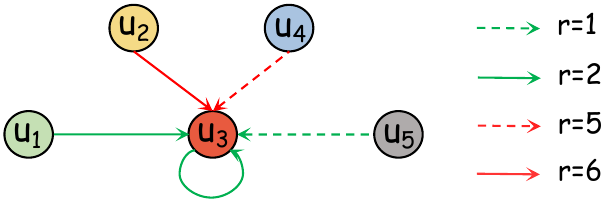}
 \caption{An example of SATG. $u_1, u_3$ and $u_5$ are from speaker 1 while $u_2$ and $u_4$ are from speaker 2.
 w.l.o.g, only the edges directed into $u_3$ node are illustrated.}
 \label{fig: sagraph}
\end{figure}

We design a SATG to model the information aggregation between utterances in a dialog.
Formally, SATG is a complete directed graph denoted as $\mathcal{G}=(\mathcal{V,E,R})$.
In this paper, the nodes in $\mathcal{G}$ are the utterances in the dialog, i.e., $\left|\mathcal{V}\right|=N,\mathcal{V}=(u_1, ..., u_N)$, and the edge $(i,j,r_{ij})\in \mathcal{E}$ denotes the information aggregation from $u_i$ to $u_j$ under the relation $r_{ij}\in \mathcal{R}$.
Table \ref{table: sagraph} lists the definitions of all relation types in $\mathcal{R}$.
In particular, there are three kinds of information conveyed by $r_{ij}$: the speaker of $u_i$, the speaker of $u_j$, and the relative position of $u_i$ and $u_j$.
Naturally, the utterances in a dialog are chronologically ordered, so the relative position of two utterances denotes their temporal relation.
An example of SATG is shown in Fig. \ref{fig: sagraph}.
Compared with the previous dialog graph structure \cite{dcrnet,cogat}, our SATG has two main advancements.
First, as a complete directed graph,
SATG can model both the intra- and inter-speaker semantic interactions. 
Second, incorporating temporal information, SATG can model the transitions of speaker-aware semantic states as the dialog goes on, which benefits both tasks.
\begin{table}[t]
\centering
\fontsize{8}{10}\selectfont
\caption{All relation types in SATG (assume there are two speakers). $I_s(i)$ indicates the speaker node $i$ is from. $pos(i,j)$ indicates the relative position of node $i$ and $j$.}
\setlength{\tabcolsep}{1.5mm}{
\begin{tabular}{c|cccccccc}
\toprule
$r_{ij}$  &1&2&3&4&5&6&7&8   \\ \midrule
$I_s(i)$ &1&1&1&1&2&2&2&2 \\
$I_s(j)$ &1&1&2&2&1&1&2&2\\
$pos(i,j)$&$>$&$\leq$&$>$&$\leq$&$>$&$\leq$&$>$&$\leq$\\
\bottomrule
\end{tabular}}
\label{table: sagraph}
\end{table}
\subsubsection{SAT-RGCN}
Inspired from \cite{rgcn}, we apply SAT-RGCN over SATG to achieve the information aggregation: 
\begin{equation}
\hat{h}_{i}=W_{1} h_{i}^0 + \sum_{r \in \mathcal{R}} \sum_{j \in \mathcal{N}_{i}^{r}} \frac{1}{\left|{N}_{i}^{r} \right|} W^r_{1} h_{j}^0
\end{equation}
where $W_{1}$ is self-transformation matrix and $W_1^r$ is relation-specific matrix.

\subsection{DTR-RSGT}
\subsubsection{Dual-task Reasoning Temporal Graph}
Inspired by \cite{dignet,kagrmn,jair,ijcaisubgraph,co-evolving}, we design a DRTG to provide an advanced platform for dual-task relational reasoning.
It is also a complete directed graph that consists of $2N$ dual nodes: $N$ sentiment nodes and $N$ act nodes.
The definitions of all relation types in $\mathcal{R'}$ are listed in Table \ref{table: dtgraph}.
Intuitively, when predicting the label of a node, the information of its dual node plays a key role, so we emphasize the temporal relation of `$=$' rather than merge it with `$<$' like SATG. 
Specifically, the relation $r'_{ij}$ conveys three kinds of information: the task of $n_i$, the task of $n_j$ and the temporal relation between $n_i$ and $n_j$.
An example of DRTG is shown in Fig. \ref{table: dtgraph}.
Compared with the previous dual-task graph structure \cite{dcrnet,cogat}, our DRTG has two major advancements.
First, the temporal relations in DRTG can make the DTR-RSGT capture the temporal information, which is essential for dual-task reasoning, while this cannot be achieved by the co-attention \cite{dcrnet} or graph attention network \cite{cogat} operating on their non-temporal graphs.
Second, in DRTG, the information aggregated into a node is decomposed by different relations that correspond to individual contributions, rather than only depending on the semantic similarity measured by the attention mechanisms.
\begin{table}[t]
\centering
\fontsize{8}{10}\selectfont
\caption{All relation types in DRTG. $I_t(i)$ indicates that node $i$ is a sentiment (S) node or act (A) node.}
\setlength{\tabcolsep}{1mm}{
\begin{tabular}{c|cccccccccccc}
\toprule
$r'_{ij}$ &1& 2& 3& 4& 5& 6& 7& 8& 9&10&11&12     \\ \midrule
$I_t(i)$          & S&S&S&S&S&S&A&A&A&A&A&A\\
$I_t(j)$          &S&S&S&A&A&A&S&S&S&A&A&A\\
$pos(i, j)$&$<$&$=$ &$>$&$<$&$=$&$>$&$<$&$=$&$>$&$<$&$=$ &$>$\\
\bottomrule
\end{tabular}}
\label{table: dtgraph}
\end{table}
\begin{figure}[t]
 \centering
 \includegraphics[width = 0.35\textwidth]{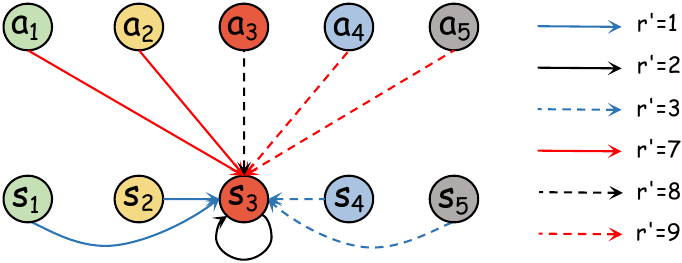}
 \caption{An example of DRTG. $s_i$ and $a_i$ respectively denote the node of DAC task and DAR task. w.l.o.g, only the edges directed into $s_3$ are illustrated.}
 \label{fig: dtgraph}
\end{figure}

\subsubsection{DTR-RSGT}
We apply DTR-RGCN to DRTG to achieve information aggregation.
Specifically, the node updating process of DTR-RGCN can be formulated as:
\begin{equation}
\overline{h}_{i}^{t}=W_{2} \hat{h}_{i}^t + \sum_{r \in \mathcal{R'}} \sum_{j \in \mathcal{N}_{i}^{r'}} \frac{1}{\left|{N}_{i}^{r'} \right|} W^r_{2} \hat{h}_{j}^t
\end{equation}
where $W_{2}$ is self-transformation matrix and $W_2^r$ is relation-specific matrix.

\section{DARER$^\textbf{2}$}
Based on the overall model introduced in Section 3, DARER$^{2}$ achieves SAT-RSGT and DTR-RSGT via applying our proposed ReTeFormers on the speaker-aware temporal graph (SATG) and dual-task reasoning temporal graph (DRTG), respectively.
Next, we first introduce the details of ReTeFormer, then the SAT-RSGT and DTR-RSGT of DARER$^2$.
\subsection{Relational Temporal Transformer}
\begin{figure*}[t]
 \centering
 \includegraphics[width = \textwidth]{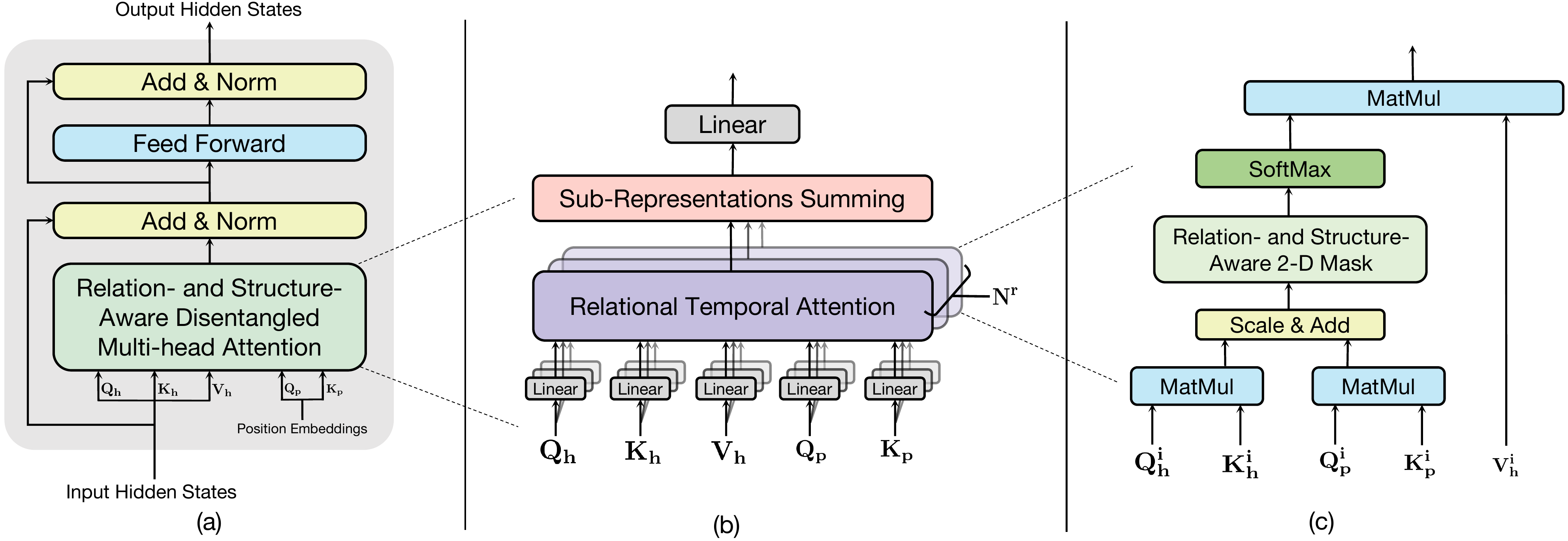}
 \caption{(a) Illustration of ReTeFormer. 
(b) Illustration of Relation- and Structure-Aware Disentangled Multi-head Attention.
(c) Illustration of Relational Temporal Attention corresponding to the $i$-th relation. $\mathbf{Q_h}, \mathbf{K_h}$ and $\mathbf{V_h}$ denote the query matrix, key matrix and value matrix of input hidden states. $\mathbf{Q_p}$ and $\mathbf{K_p}$ denote the query matrix and key matrix of the absolute position embeddings. $\mathbf{N^r}$ denotes the number of relations. 
}
 \label{fig: reteformer}
\end{figure*}
The architecture of ReTeFormer is shown in Fig. \ref{fig: reteformer}.
The core of ReTeFormer is the Relation- and Structure- Aware Disentangled Multi-head Attention, which can handle the Relation Modeling and Temporal Modeling simultaneously.

\textbf{Relational Modeling} 
The relational graph can be disentangled into different views, each of which corresponds to a specific relation and has its own adjacency matrix.
In ReTeFormer, each head of Relational Temporal Attention corresponds to a specific relation and has its own parameterization, so as to achieve the relation-specific information aggregation.
To make sure that the information aggregation of each relation is along the relation-specific structure, we design the Relation- and Structure-Aware 2-D Mask which uses the relation-specific adjacency matrix to mask the correlation matrix.
Since the final node representation receives information along multiple relations, we design the Dynamic 1-D Mask and Merge module to extract and sum each node's sub-representations obtained from multi-head Relational Temporal Attentions.

\textbf{Temporal Modeling} A dialog can be regarded as a temporal sequence of utterances. ReTeFormer utilizes position embedding to achieve temporal modeling. The position embedding is one of the foundations of Transformer \cite{transformer}, which adds the position embedding to the input representation.
However, recently it has been proven that adding together the position embeddings and word embeddings at input harms the attention and further limit the model's expressiveness because this operation brings mixed correlations between the two heterogeneous information resources (semantics and position) and unnecessary randomness in the attention \cite{rethinking}.
To this end, Ke et al. (2021) \cite{rethinking} propose to model word contextual correlation and positional correlation separately with different parameterizations and then add them together.
And our ReTeFormer follows this manner.

Next, we introduce the details of Relation- and Structure- Aware Disentangled Multi-head Attention, which is the core of our ReTeFormer\footnote{In this section, we omit the introduction of the residual connection, the layer normalization, and the feed-forward layers, whose details are the same as vanilla Transformer \cite{transformer}.}.

\subsubsection{Relation-Specific Scaled Dot-Product Attention}
For the head corresponding to relation $r$, the correlation score $\hat{\alpha}_{ij}^r$ between every two nodes is obtained via relational modeling and temporal modeling:
\begin{equation}
 \begin{aligned}
\hat{\alpha}_{i j}^r &=\frac{1}{\sqrt{d}}(Q_{[h,i]}^r) (K_{[h,j]}^r)^{T}+\frac{1}{\sqrt{d}}(Q_{[p,i]}^r)(K_{[p,j]}^{r})^{T}\\ \label{eq: mha1}
Q_{[h,i]}^r &= h_i W_Q^r, \quad K_{[h,j]}^r = h_j W_K^r\\ 
Q_{[p, i]}^r &= p_j U_Q^r, \quad K_{[p,j]}^r = p_j U_K^r 
\end{aligned}
\end{equation}
where $h_*$ $p_*$ denote the input hidden state and position embedding, respectively; $W_Q^r$ and $W_K^r$ denote the relation-specific projection matrix for the hidden states; $U_Q^r$ and $U_K^r$ denote the relation-specific projection matrix for the position embeddings; $\sqrt{d}$ is the scaling term for retaining the magnitude of $\hat{\alpha}_{i j}^r$.

Now we obtain the relation-specific correlation matrix $M^r$ for each relation $r$, which represents each two nodes' correlation along the specific relation.

\subsubsection{Relation- and Structure-Aware 2-D Mask}
Although we obtain the relation-specific correlation scores of all node pairs, a specific relation has its own adjacent structure which is crucial for information aggregation.
And the attention score between two nodes should also be calculated regarding the relation-specific neighbors.
To achieve this, we design the relation- and structure-aware 2-D mask to introduce the relation-specific structure into the attention mechanism.
Specifically, the relational graph can derive $\mathbf{N^r}$ relation-specific adjacency matrice via disentangling.
And for each relation $r$, its adjacency matrix $A^r$ is used to mask its correlation matrix $M^r$.
Finally, the normalized relation-specific attention score $\alpha_{ij}^r$ is obtained as follows:
\begin{equation}
 \begin{aligned}
\alpha_{i j}^r &= \operatorname{softmax}\left(f_{mask}^{2D}(\alpha_{ij}^r, A^r_{ij})\right)\\
f_{mask}^{2D} &= 
\begin{cases}
\hat{\alpha}^r_{ij} & A^r_{ij}=1\\
-\infty & A^r_{ij}=0\\
\end{cases}
\end{aligned}
\end{equation}
where $f_{mask}^{2D}$ denotes the function of the relation- and structure-aware 2-D mask.

\subsubsection{Output Node Representation}
For the attention head corresponding to relation $r$, the updated sub-representation of node $i$ (or the information that node $i$ should receive along relation $r$) is:
\begin{equation}
 \begin{aligned}
\hat{h}^r_i &= \sum_{j\in\mathcal{N}^r_i} \alpha_{ij} V^r_{[h,j]} \\
V^r_{[h,j]} &= h_j W_V^r
\end{aligned}
\end{equation}

A node is always connected to other nodes along different relations.
Therefore, the final updated representation node $i$ is the sum of its sub-representations of all attention heads:
\begin{equation}
\hat{h}_i = \sum_{r\in\mathcal{R}} \hat{h}^r_i
\end{equation}

\subsection{SAT-ReTeFormer}
\begin{figure}[t]
 \centering
 \includegraphics[width = 0.4\textwidth]{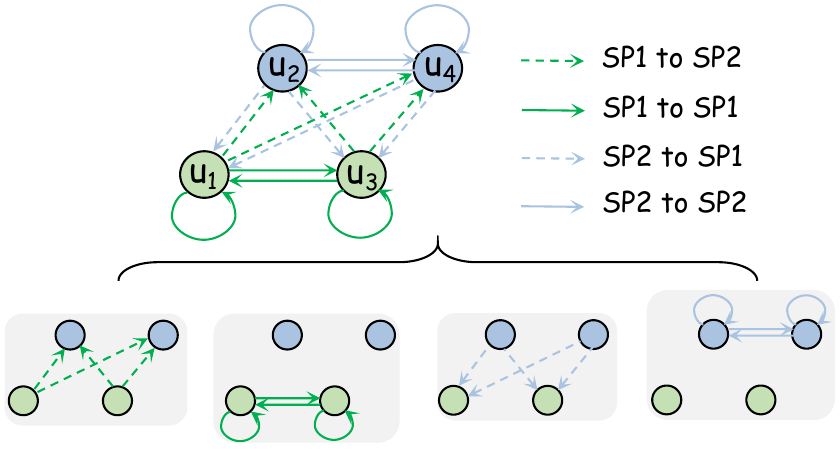}
 \caption{An example of the speaker-aware graph for SAT-ReTeFormer and its four disentangled views. Assuming there are four utterances: $u_1$ and $u_3$ (in green color) are from the speaker 1 ($SP1$); $u_2$ and $u_4$ (in blue color) are from the speaker 2 ($SP2$). Each view has its own adjacency matrix for the corresponding head of relational temporal attention.}
 \label{fig: sag}
\end{figure}

In DARER$^2$, the speaker-aware temporal RSGT (Sec. \ref{sec: sat-rsgt}) is achieved by the SAT-ReTeFormer rather than the RGCN used in DARER.
And the speaker-aware graph for SAT-ReTeFormer is shown in Fig. \ref{fig: sag}.
\begin{figure}[t]
 \centering
 \includegraphics[width = 0.37\textwidth]{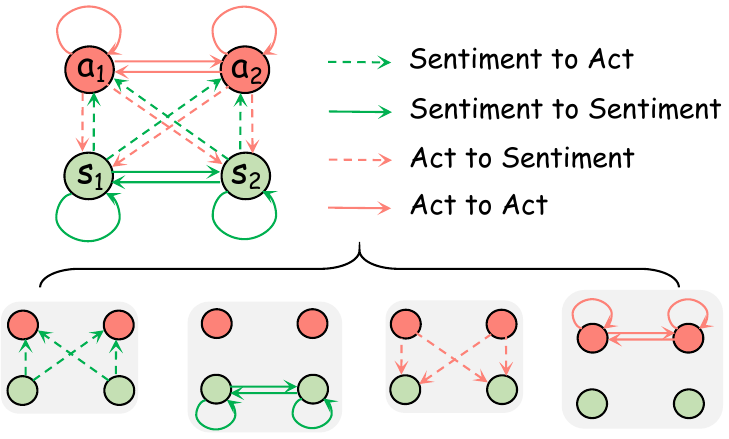}
 \caption{An example of the dual-task reasoning graph for SAT-ReTeFormer.
W.l.o.g, the dialog includes two utterances. $s_1$ and $s_2$ (in green color) denote the sentiment nodes corresponding to the first and second utterances; $a_1$ and $a_2$ (in red color) denote the act nodes of the first and second utterances.
 The graph can be disentangled into four views along the four relations, and each view has its own adjacency matrix used in the corresponding head of relational temporal attention.}
 \label{fig: drg}
\end{figure}
The input of SAT-ReTeFormer is the sequence of the initial utterance representations $H=(h_0,...,h_N)$.
In the SAT-ReTeFormer's speaker-aware graph, each node corresponds to an utterance, whose representation $h_i$ corresponds to its position embedding $p_i$.
Compared with RGCN, our SAT-ReTeFormer can explicitly model the correlations among the utterances, integrating both the speaker information and the fine-grained temporal information.
After SAT-ReTeFormer,  we obtain the context-, speaker- and temporal-sensitive utterance representations: $\hat{H}=(\hat{h_0},...,\hat{h_N})$.
\subsection{DTR-ReTeFormer}
In DARER$^2$, the dual-task reasoning RSGT (Sec. \ref{sec: dtr-rsgt}) is achieved by DTR-ReTeFormer rather than the RGCN used in DARER.
And the dual-task reasoning graph for DTR-ReTeFormer is illustrated in Fig \ref{fig: drg}.

The input of DTR-ReTeFormer is the concatenation of $\mathbf{\hat{H}_s^t}$ and $\mathbf{\hat{H}_a^t}$:
$
\mathbf{\hat{H}_s^t}\ \| \ \mathbf{\hat{H}_a^t} = [\hat{h}_{s,1}^t, ..., \hat{h}_{s,N}^t, \hat{h}_{a,1}^t, ..., \hat{h}_{a,N}^t]
$.
Since $\hat{h}_{s,i}^t$ and $\hat{h}_{a,i}^t$ corresponds to the same utterance ($u_i$), they have the same position embedding $p_i$.

Here we demonstrate the details of the semantics- and prediction-level interactions achieved by DTR-ReTeFormer.
Assuming that node $i$ is sentiment node and node $j$ is act node, the correlative score between $i$ and $j$ is calculated as follows (bringing Eq. 4 into Eq. 13):
\begin{equation}
\begin{aligned}
&\hat{\alpha}_{i j}^{r',t} =\frac{1}{\sqrt{d}}(\hat{h}^t_{s,i} W_Q^r) (\hat{h}^t_{a,j} W_K^r)^{T}+\frac{1}{\sqrt{d}}(p_j U_Q^r)(p_j U_K^r)^{T}\\
&=\frac{1}{\sqrt{d}}\left((h_{s,i}^{t-1} + e_{s,i}^t + e_{a,i}^t) W_Q^r\right) \left((h_{a,j}^{t-1} + e_{s,j}^t + e_{a,j}^t) W_K^r\right)^{T}\\ 
&\quad +\frac{1}{\sqrt{d}}(p_i U_Q^r)(p_j U_K^r)^{T}\\
&= \frac{1}{\sqrt{d}}h_{s,i}^{t-1} W_Q^r (W_K^r)^T (h_{a,j}^{t-1})^T + \frac{1}{\sqrt{d}}h_{s,i}^{t-1} W_Q^r (W_K^r)^T (e_{s,j}^{t})^T \\
& \quad + \frac{1}{\sqrt{d}} h_{s,i}^{t-1} W_Q^r (W_K^r)^T (e_{a,j}^{t})^T + \frac{1}{\sqrt{d}} e_{s,i}^{t} W_Q^r (W_K^r)^T (h_{a,j}^{t-1})^T\\
& \quad + \frac{1}{\sqrt{d}} e_{s,i}^{t} W_Q^r (W_K^r)^T (e_{s,j}^{t})^T + \frac{1}{\sqrt{d}} e_{s,i}^{t} W_Q^r (W_K^r)^T (e_{a,j}^{t})^T\\
& \quad + \frac{1}{\sqrt{d}} e_{a,i}^{t} W_Q^r (W_K^r)^T (h_{a,j}^{t-1})^T + \frac{1}{\sqrt{d}} e_{a,i}^{t} W_Q^r (W_K^r)^T (e_{s,j}^{t})^T\\
& \quad + \frac{1}{\sqrt{d}} e_{a,i}^{t} W_Q^r (W_K^r)^T (e_{a,j}^{t})^T + \frac{1}{\sqrt{d}}(p_j U_Q^r)(p_j U_K^r)^{T}
\end{aligned}
\end{equation}
In $\hat{\alpha}_{i j}^{r',t}$, $r'$ denotes the relation of \textit{Act to Sentiment} and $t$ denotes the time step of recurrent dual-task reasoning.
We can observe that finally, there are 10 terms in Eq. 17.
The 1st term models the semantics-level interaction between node $i$ and node $j$.
The 2nd-9th terms model the prediction-level interactions.
Specifically, the 2nd, 3rd, 4th and 7th terms model the semantics-prediction interactions;
the 5th, 6th and 9th terms model the prediction-prediction interactions.
The 10th term achieves relation-specific temporal modeling.

Therefore, our proposed DTR-ReTeFormer can explicitly and comprehensively model the semantics- and prediction-level interactions.
And relational modeling achieves the self-task and cross-task interactions.
Besides, in this process, the relation-specific temporal information is considered, facilitating dual-task reasoning.

\section{Experiments}
\subsection{Datasets and Metrics}
\textbf{Datasets}. We conduct experiments on two publicly available dialogue datasets: Mastodon
\cite{mastodon} and Dailydialog
\cite{dailydialog}.
The Mastodon dataset includes 269 dialogues for training and 266 dialogues for testing. 
And there are 3 sentiment classes and 15 act classes.
Since there is no official validation set, we follow the same partition as \cite{cogat}.
Finally, there are 243 dialogues for training, 26 dialogues for validating, and 266 dialogues for testing.
As for Dailydialog dataset, we adopt the official train/valid/test/ split from the original dataset \cite{dailydialog}: 11,118 dialogues for training, 1,000 for validating, and 1,000 for testing. And there are 7 sentiment classes and 4 act classes.\\
\textbf{Evaluation Metrics}. Following previous works \cite{mastodon,dcrnet,cogat}, on Dailydialog dataset, we adopt macro-average Precision (P), Recall (R), and F1 for the two tasks, while on Mastodon dataset, we ignore the neutral sentiment label in DSC task and for DAR task we adopt the average of the F1 scores weighted by the prevalence of each dialogue act.

\begin{table*}[t]
\centering
\fontsize{8}{9}\selectfont
\caption{Experiment results. $^*$ denotes we reproduce the results using official code. $^\dag$ denotes that our DARER and DARER$^2$ significantly outperforms the previous best model Co-GAT with $p<0.01$ under t-test and $^\ddag$ denotes $p<0.05$. $\uparrow$ denotes the improvement achieved by our model over Co-GAT.}
\setlength{\tabcolsep}{1.6mm}{
\begin{tabular}{c|ccc|ccc|ccc|ccc}
\toprule
\multirow{3}{*}{Models} & \multicolumn{6}{c|}{Mastodon}                  & \multicolumn{6}{c}{DailyDialog}   \\ \cline{2-13}
            & \multicolumn{3}{c|}{DSC} & \multicolumn{3}{c|}{DAR} & \multicolumn{3}{c|}{DSC} & \multicolumn{3}{c}{DAR}\\\cline{2-13}
                        & P(\%)     & R(\%) & F1(\%) & P(\%) & R(\%)  & F1(\%)& P(\%)  & R(\%)  & F1(\%)& P(\%) & R(\%) & F1(\%)\\ \midrule
    JointDAS \cite{mastodon}            &36.1   & 41.6  & 37.6  & 55.6  & 51.9  & 53.2  & 35.4  & 28.8  & 31.2  & 76.2  & 74.5  & 75.1  \\ 
     IIIM \cite{kimkim}   & 38.7  & 40.1  & 39.4  & 56.3  & 52.2  & 54.3  & 38.9  & 28.5  & 33.0  & 76.5  & 74.9  & 75.7  \\  
  DCR-Net \cite{dcrnet}& 43.2  & 47.3  & 45.1  & 60.3  & 56.9  & 58.6  & 56.0  & 40.1  & 45.4  & 79.1  & 79.0  & 79.1  \\  
  BCDCN \cite{bcdcn} &38.2 &62.0 &45.9 &57.3 &61.7 &59.4 &55.2 &45.7 &48.6 &80.0 &80.6 &80.3\\ 
   Co-GAT \cite{cogat} & 44.0  & 53.2  & 48.1  & 60.4  & 60.6  & 60.5  & 65.9  & 45.3  & 51.0  & 81.0  & 78.1  & 79.4  \\ 
 {Co-GAT$^*$}& 45.40 & 48.11 & 46.47 & 62.55 & 58.66 & 60.54 & 58.04 & 44.65 & 48.82 & 79.14 & 79.71 & 79.39 \\
\midrule
 \multirow{2}{*}{DARER} & \textbf{56.04}$^\dag$ & \textbf{63.33}$^\dag$ & \textbf{59.59}$^\dag$ & \textbf{65.08}$^\ddag$ & \textbf{61.88}$^\dag$ & \textbf{63.43}$^\dag$ & \textbf{59.96}$^\ddag$ & \textbf{49.51}$^\dag$ & \textbf{53.42}$^\dag$ & \textbf{81.39}$^\dag$ & \textbf{80.80}$^\ddag$ & \textbf{81.06}$^\dag$ \\ 
 & $\uparrow$23.4\% & $\uparrow$31.6\% & $\uparrow$28.2\% & $\uparrow$4.0\% & $\uparrow$5.5\% & $\uparrow$4.8\% &$\uparrow$3.3\% & $\uparrow$10.9\% & $\uparrow$9.4\% & $\uparrow$2.8\%&$\uparrow$1.4\% &$\uparrow$2.1\% \\ \hline
 \multirow{2}{*}{DARER$^2$} & \textbf{58.53}$^\dag$ & \textbf{67.06}$^\dag$ & \textbf{62.38}$^\dag$ & \textbf{68.26}$^\dag$ & \textbf{67.15}$^\dag$ & \textbf{67.70}$^\dag$ & \textbf{65.58}$^\dag$ & \textbf{48.28}$^\dag$ & \textbf{54.34}$^\dag$ & \textbf{81.41}$^\dag$ & \textbf{81.79}$^\ddag$ & \textbf{81.60}$^\dag$ \\ 
 & $\uparrow$28.9\% & $\uparrow$39.4\% & $\uparrow$34.2\% & $\uparrow$9.1\% & $\uparrow$14.5\% & $\uparrow$11.8\% &$\uparrow$13.0\% & $\uparrow$8.1\% & $\uparrow$11.3\% & $\uparrow$2.9\%&$\uparrow$2.6\% &$\uparrow$2.8\% \\ 
 \bottomrule
\end{tabular}}
\label{table: results}
\end{table*}

\subsection{Implement Details and Baselines}
Both of DARER and DARER$^2$ are trained with Adam optimizer with the learning rate of $1e^{-3}$ and the batch size is 16.
We exploit 300-dimensional Glove vectors for the word embeddings.
And the epoch number is 100 for Mastodon and 50 for DailyDialog.
Next, we introduce the different settings of other hyper-parameters for DARER and DARER$^2$

\textbf{DARER}
The dimension of hidden states (label embeddings) is 128 for Mastodon and 256 for DailyDialog. 
The step number $T$ for recurrent dual-task reasoning is set to 3 for Mastodon and 1 for DailyDialog.
The coefficient $\gamma_s$ and $\gamma_a$ are 3 for Mastodon and $1e^{-4}$ for DailyDialog.
To alleviate overfitting, we adopt dropout, and the ratio is 0.2 for Mastodon and 0.3 for DailyDialog.

\textbf{DARER$^\textbf{2}$}
The dimension of hidden states (label embeddings) is 256 for Mastodon and 300 for DailyDialog. 
The step number $T$ for recurrent dual-task reasoning is set to 5 for Mastodon and 3 for DailyDialog.
For Mastodon dataset, the coefficients $\gamma_s$ and $\gamma_a$ are 10 and 1.
For DailyDialog, the coefficients $\gamma_s$ and $\gamma_a$ are 0.1 and $1e^{-6}$.
The dropout ratio is 0.4 for both Mastodon and DailyDialog.

For all experiments, we pick the model performing best on the validation set and then report the average results on the test set based on three runs with different random seeds.
All computations are conducted on NVIDIA RTX 6000.

We compare our model with: JointDAS \cite{mastodon}, IIIM \cite{kimkim}, DCR-Net (Co-Attention) \cite{dcrnet}, BCDCN \cite{bcdcn} and Co-GAT \cite{cogat}.

\subsection{Main Results} \label{sec: mainresult}

\subsubsection{Comparison with Baselines}
\begin{figure}[t]
 \centering
 \includegraphics[width = 0.48\textwidth]{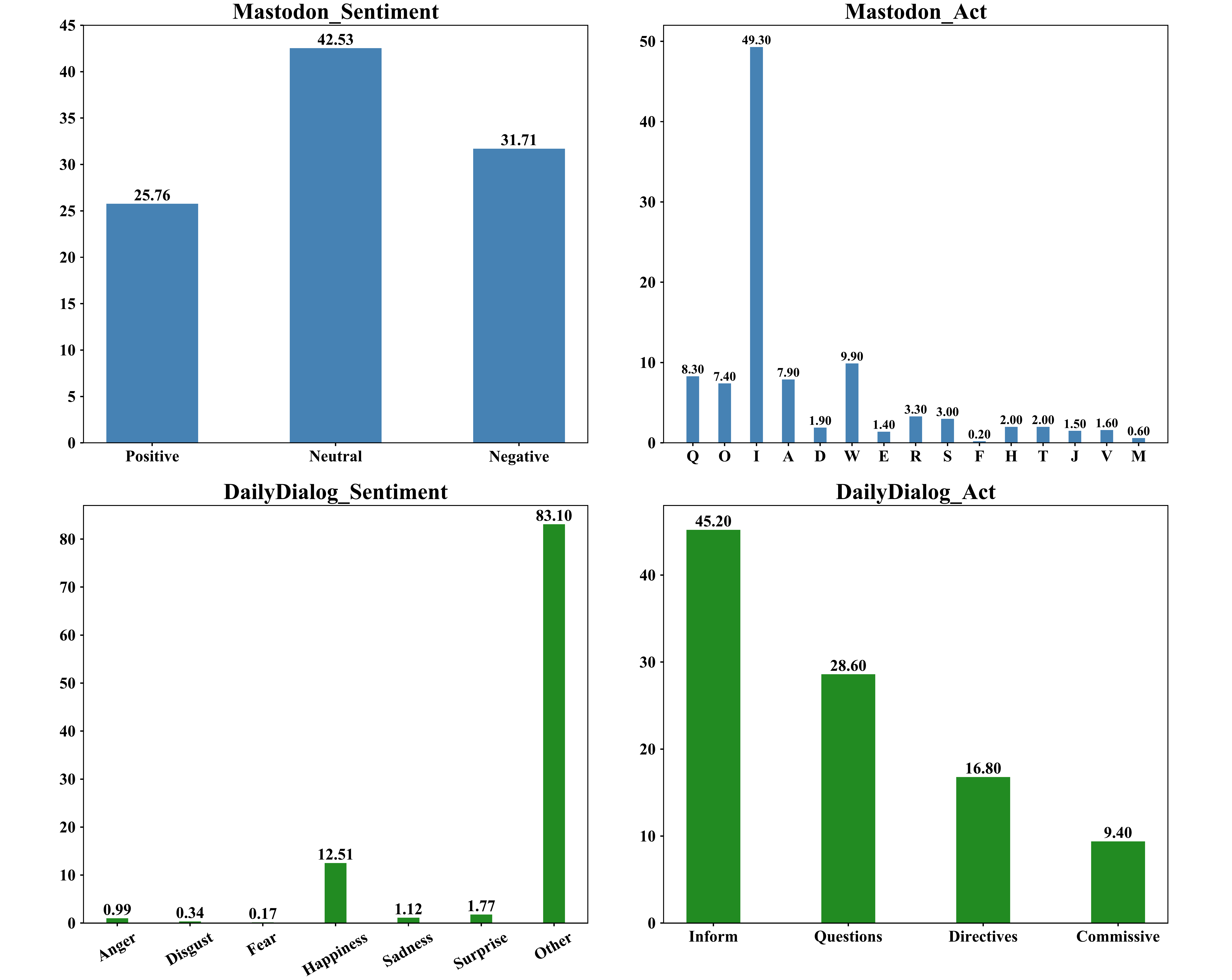}
 \caption{Illustration of class distributions on Mastodon and DailyDialog datasets.}
 \label{fig: label_distrib}
\end{figure}

Table \ref{table: results} lists the experiment results on the test sets of the two datasets.
We can observe that:\\
1. Our models significantly outperform all baselines, achieving new state-of-the-art (SOTA).
In particular, over Co-GAT, the existing SOTA, DARER achieves an absolute improvement of 13.1\% in F1 score on DSC task in Mastodon, a relative improvement of over 28\%.
And DARER$^2$ achieves even larger improvements: over 34\% improvement in F1 score on DSC task in Mastodon dataset.
The satisfying results of DARERs come from 
(1) our framework integrates not only semantics-level interactions but also prediction-level interactions, thus capturing explicit dependencies other than implicit dependencies; 
(2) our SATG represents the speaker-aware semantic states transitions, capturing the important basic semantics benefiting both tasks;
(3) our DRTG provides a rational platform on which more effective dual-task relational reasoning is conducted.
(4) the advanced architecture of our DARER models allows DSC and DAR to improve each other in the recurrent dual-task reasoning process gradually. \\
2. DARER and DARER$^2$ show more prominent superiority on DSC task than DAR task.
We surmise the probable reason is that generally, the act label is more complicated to deduce than the sentiment label in dual-task reasoning.
For instance, it is easy to infer $u_i$'s Negative label on DSC given $u_{i}$'s Agreement label on DAR and $u_{i-1}$'s Negative label on DSC.
Reversely, given the label information that $u_i$ and $u_{i-1}$ are both negative on DSC, it is hard to infer the act label of $u_i$ because there are several act labels possibly satisfying this case, e.g., Disagreement, Agreement, Statement.\\
3. Our models' improvements on DailyDialog are smaller than those on Mastodon.
We speculate this is caused by the extremely unbalanced sentiment class distribution in DailyDialog.
As shown in Fig. \ref{fig: label_distrib}, in DailyDialog dataset, over 83\% utterances do not express sentiment, while the act labels are rich and varied.
This hinders DARER from learning valuable correlations between the two tasks.
\begin{figure}[t]
 \centering
 \includegraphics[width = 0.48\textwidth]{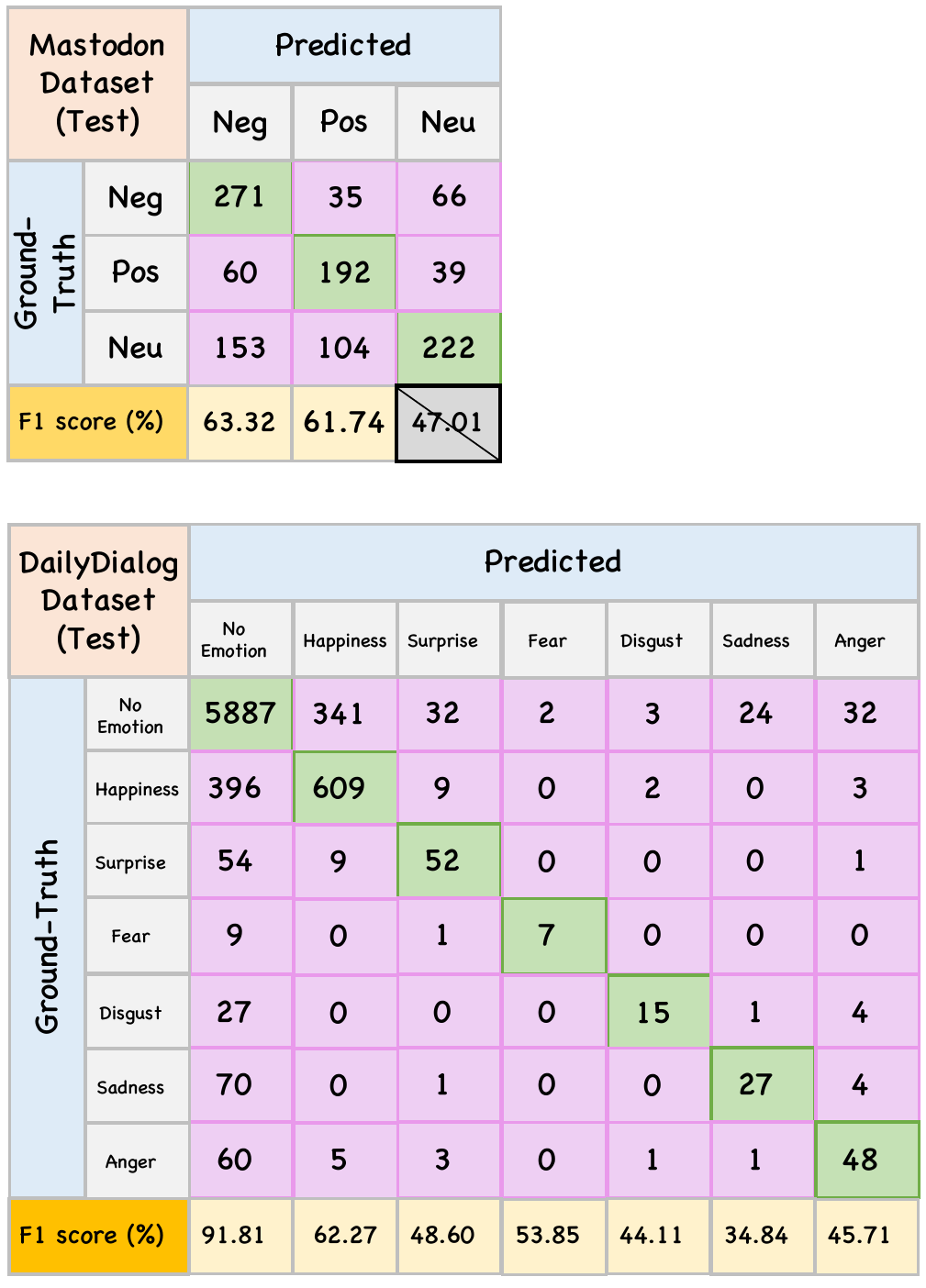}
 \caption{Illustration of confusion matrices and F1 score on each class on Mastodon and DailyDialog test sets. Note that on Mastodon test set, following previous works, the F1 score of the Neutral class is not counted for the final F1 score.}
 \label{fig: conf_mat}
\end{figure}
\subsubsection{Comparison of DARER and DARER$^2$}
From Table \ref{table: results}, we can find that DARER$^2$ outperforms DARER, further improving the performance.
This can be attributed to the fact that the proposed SAT-ReTeFormer and DTR-ReTeFormer in DARER$^2$ can more effectively model the relational and temporal interactions than the RGCNs adopted in DARER.
Especially, in DTR-ReTeFormer, since the input hidden state is superimposed with both tasks' label representation of the corresponding utterance, the relation- and structure-aware disentangled multi-head attention can explicitly and sufficiently model the relation-specific dual-task interactions, including semantics-semantics interactions, semantics-prediction interactions, and prediction-prediction interactions.

\subsubsection{Effect of Pre-trained Language Model} 
\label{sec: ptlm}
\begin{table}[t]
\centering
\fontsize{8}{10}\selectfont
\caption{Results comparison based on different PTLM encoders.}
\setlength{\tabcolsep}{1.3mm}{
\begin{tabular}{c|l|ccc|ccc}
\toprule
\multicolumn{2}{c|}{\multirow{3}{*}{\quad Models}} & \multicolumn{6}{c}{Mastodon}    \\ \cline{3-8}
         \multicolumn{2}{c|}{}   & \multicolumn{3}{c|}{DSC} & \multicolumn{3}{c}{DAR}\\ \cline{3-8}
                   \multicolumn{2}{c|}{}     & P(\%)     & R(\%) & F1(\%)    & P(\%)     & R(\%) & F1(\%)   \\ \midrule
\multirow{4}{*}{\rotatebox{90}{BERT}}
& + Linear         & 61.79 & 61.09 & 60.60 & 70.20 & 67.49 & 68.82 \\
& + Co-GAT           & 66.03 & 58.13 & 61.56 & 70.66 & 67.62 & 69.08 \\ \cline{2-8}
& + DARER         & {65.98} & {67.39} & \textbf{66.42} & {73.82} & {71.67} & \textbf{72.73} \\ 
& + DARER$^2$         & {64.47} & {71.10} & \textbf{67.61} & {75.34} & {73.04} & \textbf{74.17} \\ \midrule
\multirow{4}{*}{\rotatebox{90}{RoBERTa}}
&+ Linear     & 57.83 & 60.54 & 57.83 & 62.49 & 61.93 & 62.20 \\
& + Co-GAT        & 61.28 &57.25 & 58.26  &66.46 & 64.01 & 65.21 \\  \cline{2-8}
& + DARER      & {61.36} & {67.27} & \textbf{63.66} & {70.87} & {68.68} & \textbf{69.75} \\ 
& + DARER$^2$         & {63.78} & {71.44} & \textbf{66.49} & {73.86} & {72.87} & \textbf{73.36} \\ \midrule
\multirow{4}{*}{\rotatebox{90}{XLNet}}    
&+ Linear   & 61.42 & 67.80 & 63.35 & 67.31 & 63.04 & 65.09 \\
& + Co-GAT           & 64.01 & 65.30 & 63.71 & 67.19 & 64.09 & 65.60 \\  \cline{2-8}
& + DARER         & {68.05} & {69.47} & \textbf{68.66} & {72.04} & {69.63} & \textbf{70.81} \\ 
& + DARER$^2$         & {67.20} & {74.24} & \textbf{70.42} & {72.45} & {71.47} & \textbf{71.96} \\ \bottomrule
\end{tabular}}
\label{table:ptlm}
\end{table}
In this section, we study the effects of three PTLM encoders: BERT \cite{bert}, RoBERTa \cite{roberta}, and XLNet \cite{xlnet}, which replace the BiLSTM utterance encoder in the state-of-the-art model Co-GAT and our DARER models.
We adopt the base versions of the PTLMs implemented in PyTorch by \cite{transformers}.
In our experiments, the whole models are trained by AdamW optimizer with the learning rate of $1e^{-5}$ and the batch size is 16.
And the PTLMs are fine-tuned in the training process.
Results are listed in Table \ref{table:ptlm}.
We can find that since single PTLM encoders are powerful in language understanding, they obtain promising results even without any interactions between utterances or the two tasks.
Nevertheless, stacking DARER on PTLM encoders further obtains around 4\%-10\% absolute improvements on F1.
This is because our DARER models achieve relational temporal graph reasoning prediction-level interactions, which complement the high-quality semantics grasped by PTLM encoders.
In contrast, Co-GAT only models the semantics-level interactions, whose advantages are diluted by the powerful PTLMs.
Consequently, based on PTLM encoders, Co-GAT brings much less improvement than our DARER models.

\subsection{Ablation Study}
\begin{table}[t]
\centering
\fontsize{8}{10}\selectfont
\caption{Results (in F1 score) of ablation experiments on DARER.} 
\setlength{\tabcolsep}{1.6mm}{
\begin{tabular}{c|cc|cc}
\toprule
\multirow{2}{*}{Variants} & \multicolumn{2}{c|}{Mastodon} & \multicolumn{2}{c}{DailyDialog} \\ \cline{2-5} 
                        & DSC             & DAR            & DSC              & DAR             \\ \midrule
        DARER         & \textbf{59.59}       &   \textbf{63.43}       &  \textbf{53.42}        &  \textbf{81.06}        \\ \hline
  w/o Label Embeddings   & 56.76        &   62.15       & 50.64         &  79.87              \\ 
  w/o $\mathcal{L}_{constraint}$               & 56.22         & 61.99      &   49.94        &   79.76        \\ \hline
     w/o SAT-RSGT        & 57.37        &  62.96      &  50.25        &  80.52        \\ 
     w/o DTR-RSGT       & 56.69       &   61.69       &  50.11          & 79.76          \\
 w/o TS-LSTMs       & 56.30      &   61.49      &   51.61         &  80.33 \\ \hline 
w/o Tpl Rels in SATG  & 58.23         &  62.21       &  50.99        &  80.70      \\ 
w/o Tpl Rels in DRTG  & 57.22       &   62.15      &   50.52        &  80.28     \\ 
  \bottomrule
\end{tabular}}
\label{table: ablation}
\end{table}
Our DARER and DARER$^2$ share the same overall model architectures, which include \textbf{label embeddings}, \textbf{constraint loss}, \textbf{SAT-RSGT}, \textbf{DTR-RSGT}, \textbf{TS-LSTMs}, \textbf{SATG} and \textbf{DRTG}.
To study the effectiveness of each component, we conduct ablation experiments on DARER and Table \ref{table: ablation} lists the results.

\begin{figure*}[t]
 \centering
 \includegraphics[width = 0.8\textwidth]{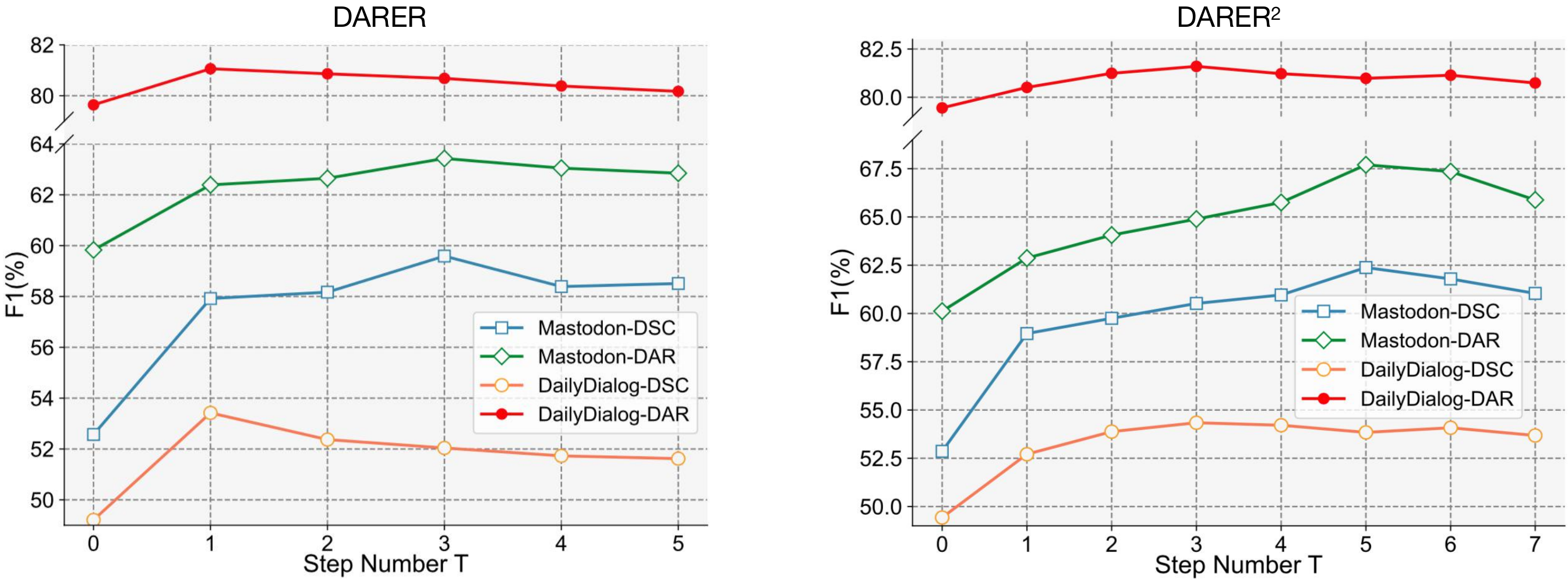}
 \caption{Performances of DARER and DARER$^2$ over different $T$.}
 \label{fig: T}
\end{figure*}

\begin{table}[t]
\centering
\caption{The comparison of DARER$^2$ and $\mathcal{L}_{constraint}$ on the performances of each single step.}
\fontsize{8}{10}\selectfont
\setlength{\tabcolsep}{1.2mm}{
\begin{tabular}{c|c|cccccc}
\hline
\multirow{2}{*}{Model}         & \multirow{2}{*}{metric} & \multicolumn{6}{c}{step}                                                                                                                              \\ \cline{3-8} 
                               &                         & \multicolumn{1}{c|}{0}     & \multicolumn{1}{c|}{1}     & \multicolumn{1}{c|}{2}     & \multicolumn{1}{c|}{3}     & \multicolumn{1}{c|}{4}     & 5     \\ \hline
\multirow{3}{*}{DARER$^2$}        & F1                      & \multicolumn{1}{c|}{57.34} & \multicolumn{1}{c|}{59.41} & \multicolumn{1}{c|}{61.70} & \multicolumn{1}{c|}{62.30} & \multicolumn{1}{c|}{62.38} & 62.53 \\ \cline{2-8} 
                               & P                       & \multicolumn{1}{c|}{59.24} & \multicolumn{1}{c|}{59.94} & \multicolumn{1}{c|}{59.35} & \multicolumn{1}{c|}{58.08} & \multicolumn{1}{c|}{57.48} & 57.00 \\ \cline{2-8} 
                               & R                       & \multicolumn{1}{c|}{55.65} & \multicolumn{1}{c|}{59.29} & \multicolumn{1}{c|}{64.96} & \multicolumn{1}{c|}{67.19} & \multicolumn{1}{c|}{68.36} & 69.41 \\ \hline
\multirow{3}{*}{w/o $\mathcal{L}_{constraint}$} & F1                      & \multicolumn{1}{c|}{54.15} & \multicolumn{1}{c|}{58.66} & \multicolumn{1}{c|}{58.89} & \multicolumn{1}{c|}{59.51} & \multicolumn{1}{c|}{58.54} & 58.75 \\ \cline{2-8} 
                               & P                       & \multicolumn{1}{c|}{58.56} & \multicolumn{1}{c|}{62.04} & \multicolumn{1}{c|}{55.33} & \multicolumn{1}{c|}{57.83} & \multicolumn{1}{c|}{55.57} & 55.46 \\ \cline{2-8} 
                               & R                       & \multicolumn{1}{c|}{51.73} & \multicolumn{1}{c|}{62.04} & \multicolumn{1}{c|}{62.95} & \multicolumn{1}{c|}{61.36} & \multicolumn{1}{c|}{61.88} & 62.45 \\ \hline
\end{tabular}}
\label{table: step predict}
\end{table}

From Table  \ref{table: ablation}, we have the following observations:\\
(1)
Removing \textbf{label embeddings} causes prediction-level interactions not to be achieved.
The sharp drops in results prove that our method of leveraging label information to achieve prediction-level interactions effectively improves dual-task reasoning via capturing explicit dependencies.\\
(2) Without \textbf{constraint loss}, the two logic rules can hardly be met, so there is no constraint forcing DSC and DAR to gradually prompt each other, resulting in the dramatic decline of performances. 
(3) As the core of Dialog Understanding, \textbf{SAT-RSGT} captures speaker-aware semantic states transitions, which provides essential basic task-free knowledge for both tasks.
Without it, some essential indicative semantics would be lost, then the results decrease.\\
(4) The worst results of `w/o DTR-RSGT' prove that \textbf{DTR-RSGT} is the core of DARER, and it plays a vital role in conducting dual-task relational reasoning over the semantics and label information.\\
(5) The significant results decrease of `w/o TS-LSTMs' prove that \textbf{TS-LSTMs} also plays an important role in DARER by generating task-specific hidden states for both tasks and have some capability of sequence label-aware reasoning.\\
(6) Removing of the \textbf{temporal relations} (Tpl Rels) in SATG or DRTG causes distinct results decline.
This can prove the necessity and effectiveness of introducing temporal relations into dialog understanding and dual-task reasoning.

To further study the necessity of $\mathcal{L}_{constraint}$, we compare DARER$^2$ and the variant w/o $\mathcal{L}_{constraint}$ on the detailed performances of each single step, as shown in Table \ref{table: step predict}. We can observe that the $\mathcal{L}_{constraint}$ in DARER2 can make the model generate better predictions at each step than w/o $\mathcal{L}_{constraint}$. Besides, thanks to the margin loss, DARER$^2$ can generate better and better predictions along the time step. However, removing $\mathcal{L}_{constraint}$ leads to fluctuations and significant drops in performance. The reason is that without the estimate loss and margin loss, the two rules cannot be integrated into the training process. Only relying on the cross-entropy loss at the final step cannot effectively improve the predictions of the previous step nor make the model generate better and better predictions along the step.

\subsection{Superiority of ReTeFormer}
\begin{table}[t]
\centering
\fontsize{8}{10}\selectfont
\caption{Results (in F1 score) of different settings of ReTeFormers. $\times$ denotes the corresponding ReTeFormer is replaced with the RGCN counterpart used in DARER.} 
\setlength{\tabcolsep}{2.5mm}{
\begin{tabular}{c|c|c||cc|cc}
\toprule
\multirow{2}{*}{Variants} & \multicolumn{2}{c||}{-ReTeFormer} & \multicolumn{2}{c|}{Mastodon} & \multicolumn{2}{c}{DailyDialog} \\ \cline{2-7} 
        &SAT   &DTR                & DSC             & DAR      & DSC         & DAR         \\ \midrule
DARER$^2$ & $\checkmark$ & $\checkmark$ & \textbf{62.38} &  \textbf{67.70} &  \textbf{54.34} & \textbf{81.60} \\ \hline
M$_1$ & $\times$ & $\checkmark$ & 61.85 &  66.57 &  54.16 & 81.54 \\ 
M$_2$ & $\checkmark$ & $\times$ & 60.25 &  64.23 &  53.98 & 81.22 \\ \hline
DARER & $\times$ & $\times$  & 59.59       &   63.43       &  {53.42}        &  {81.06}        \\
  \bottomrule
\end{tabular}}
\label{table: reteformer}
\end{table}
In DARER$^2$ the SAT-RSGT and DTR-RSGT are achieved by our proposed SAT-ReTeFormer and DTR-ReTeFormer respectively, rather than RGCNs.
To verify the superiorities of the two ReTeFormers over their RGCN counterparts in DARER, we change the setting of the two ReTeFormers and show the performances in Table \ref{table: reteformer}.

We can observe that for both SATG and DTRG, our ReTeFomer shows significant superiority over RGCN.
There are two main reasons.
First, our proposed ReTeFormer can conduct \textit{fine-grained} relational temporal modeling, while RGCN can only handle the \textit{coarse-grained} relative temporal relations.
Intuitively, fine-grained relational temporal modeling can better model the latent structures of the dialog than coarse-grained one, further benefiting dual-task reasoning.
Second, the Relation- and Structure-Aware Disentangled Multi-head Attention in our proposed ReTeFormer can explicitly model the correlations between the nodes.
Especially, our DTR-ReTeFormer can explicitly and comprehensively model the self-task and cross-task interactions that are of both semantics- and prediction-level.

\subsection{Impact of Step Number $T$}
The performances of DARER and DARER$^2$ over different $T$ are plotted in Fig. \ref{fig: T}.
$T=0$ denotes the output of the Initial Estimation module is regarded as final predictions.
We can find that appropriately increasing $T$ brings results improvements.
Particularly, with $T$ increasing from 0 to 1, the results increase sharply.
This verifies that the Initial Estimation module can provide useful label information for dual-task reasoning.
Furthermore, DARER can learn beneficial mutual knowledge from recurrent dual-task reasoning in which DSC and DAR prompt each other.
Generally, when $T$ surpasses a certain point, the performances decline slightly.
The possible reason is that after the peak, more dual-task interactions cause too much deep information fusion of the two tasks, leading to the loss of some important task-specific features and overfitting.

\subsection{Case Study}
\begin{figure*}[t]
 \centering
 \includegraphics[width = 0.8\textwidth]{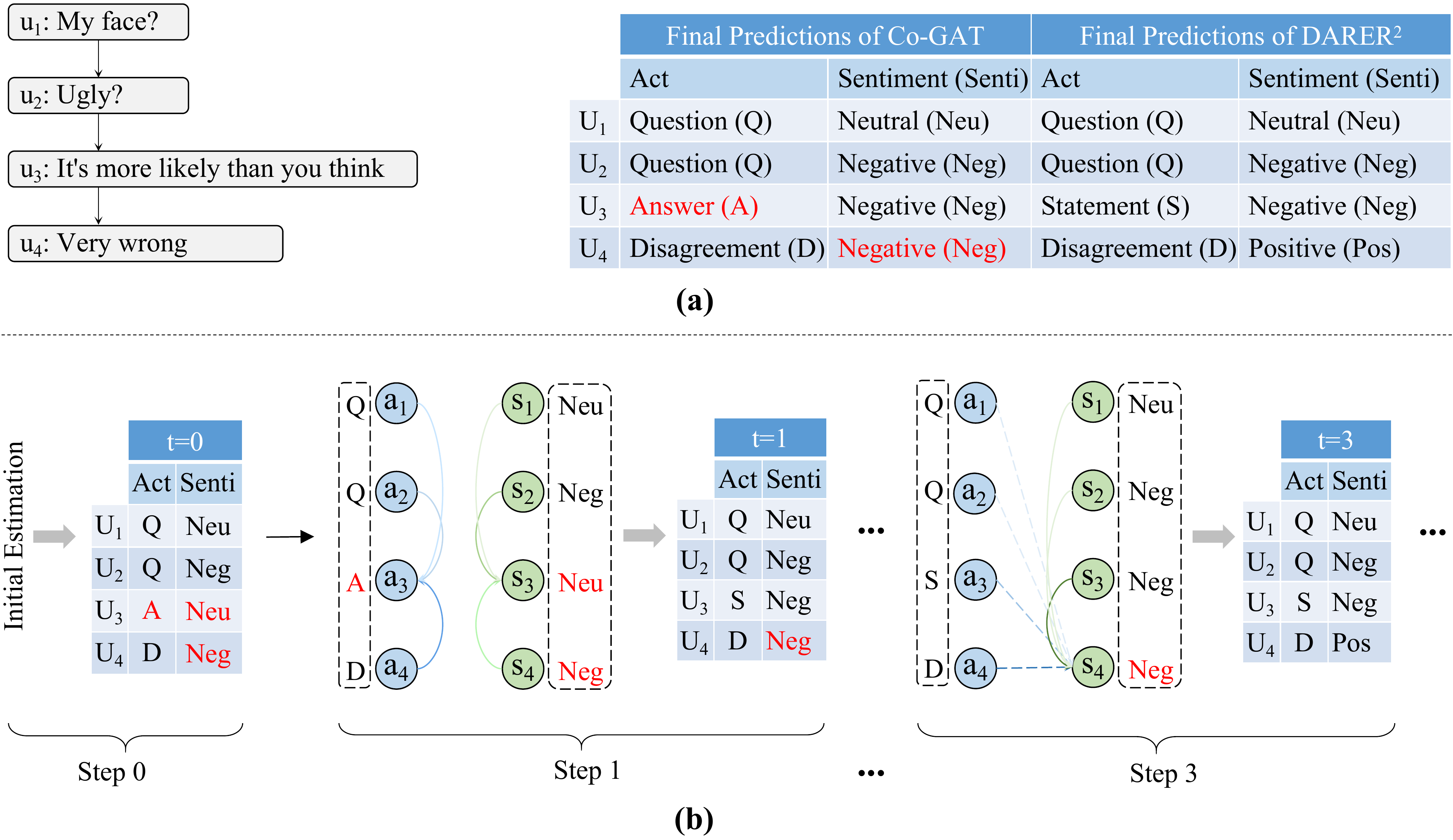}
 \caption{Case study. (a) The example dialog and the final predictions of Co-GAT and our DARER$^2$. The red color denotes error. (b) Illustration of the estimated labels at each time step and the reasoning process. For simplification, we only list the highest probability label rather than the whole label distribution. The dashed box denotes the label estimated at the previous step. $a_i$ and $s_i$ denote the act node and sentiment node of $u_i$, respectively. The blue solid arrows denote the edges between act nodes. The green solid arrows denote the edges between sentiment nodes. The blue dashed arrows denote the edges from act nodes to sentiment nodes. Deeper color denotes a larger attention weight.}
 \label{fig: case}
\end{figure*}
To better understand how our model works well, we compare the final predictions of Co-GAT and our DARER$^2$, as shown in Fig. \ref{fig: case} (a). 
We can find that our DARER$^2$ can correctly predict all labels of both tasks, while there are some errors in Co-GAT's predictions: the act label of $u_3$ is incorrectly inferred as \texttt{Answer}, and the sentiment label of $u_4$ is incorrectly inferred as \texttt{Negative}.
We suppose there are two reasons: (1) Co-GAT works on a homogeneous fully-connected dual-task graph, losing the intra- and cross-task dependencies and temporal information among the nodes; (2) Co-GAT only achieves semantics-level to implicitly models the dual-task dependencies, without incorporating prediction-level interactions.

To show how our DARER$^2$ conducts the dual-task relational temporal graph reasoning, we illustrate the dual-task reasoning process in Fig. \ref{fig: case} (b).
At step 0, the initial estimation module produces the initial label distributions.
In the first step of dual-task reasoning, some errors in the previously estimated labels are corrected through the intra-task interactions of act recognition task and sentiment classification task.
Specifically, in act nodes, the semantics and label information of node $a_4$ is assigned a large weight and aggregated into node $a_3$.
The \texttt{Disagreement} label of $a_4$ can indicate the \texttt{Statement} label of $a_3$.
This is because in the dataset, if an utterance has a \texttt{Disagreement} act label, in most cases, its previous utterance has a \texttt{Statement} act label, which is also consistent with the real-world scenarios.
In sentiment nodes, $s_2$ is assigned large weight and aggregated into node $s_3$.
Combining the semantics `more' of $s_3$ and the \texttt{Negative} label of $s_2$, the \texttt{Negative} label of $s_3$ can be correctly inferred.
Then in the third step of dual-task reasoning, the wrong label of $s_4$ node is fixed.
Specifically, node $a_3$, $a_4$ and $s_3$ are assigned relatively large attention weights for $s_4$.
Regarding the labels of $a_3$ and $a_4$, $u_4$ disagrees $u_3$, indicating $s_3$ and $s_4$ may have opposite sentiments.
And further considering the \texttt{Negative} label of $s_4$, our model can produce the correct \texttt{Positive} sentiment label for $u_4$.
In this way, our model can gradually generate better labels through recurrent relational temporal graph reasoning.

\subsection{Computation Efficiency}
\begin{table}[t]
\centering
\fontsize{8}{10}\selectfont
\caption{Comparison with SOTA on model parameters, training time, GPU memory required, and performance. }
\setlength{\tabcolsep}{1.6mm}{
\begin{tabular}{c|c|c|c|c}
\toprule
Models  & \begin{tabular}[c]{@{}l@{}}Number of \\ Parameters $\downarrow$\end{tabular} & \begin{tabular}[c]{@{}l@{}}Training Time\\ per Epoch $\downarrow$\end{tabular} & \begin{tabular}[c]{@{}l@{}}GPU $\downarrow$\\ Memory \end{tabular}  & Avg. F1$\uparrow$  \\ \midrule
Co-GAT  &3.61M        & 1.86s   &1531MB  &  53.66\% \\ \hline\hline
DARER & 2.50M       & 1.81s  & 1187MB &  61.51\%\\ \hline 
\textbf{Improve} & \textbf{-30.7}\%   &    \textbf{-2.7}\% &\textbf{-22.4}\%   & \textbf{+14.6}\%  \\ \midrule
DARER$^2$ & 3.83M       & 2.16s  & 1191MB &  65.04\%  \\ \hline 
\textbf{Improve} & +6.1\%   & +16.1\% &\textbf{-22.2}\%   & \textbf{+21.2}\%  \\ \bottomrule

\end{tabular}}
\label{table: computing_efficient}
\end{table}
In practical application, in addition to the performance, the number of parameters, the time cost, and GPU memory required are important factors.
Taking Mastodon as the testbed, we compare our DARER models with the up-to-date SOTA (Co-GAT) on these factors, and the results are shown in Table \ref{table: computing_efficient}. Avg. F1 denotes the average of the F1 scores on the two tasks.
We can find that although our DARER models surpass SOTA by a large margin, they do not significantly cost more computation resources.
Especially, DARER is even more efficient than Co-GAT.
As for DARER$^2$, although it has some more parameters and costs more training time, this is acceptable considering that it can save about 22\% GPU memory and improve 21\% performance.
Therefore, our DARER models are relatively efficient for practical application.

\subsection{Experiment on joint Multiple Intent Detection and Slot Filling }
To verify the generality of our method, we further conduct experiments on the task of joint multiple intent detection and slot filling.

\subsubsection{Task Definition}
The input is an utterance that can be denoted as $U=\{u_i\}^n_1$.
Multiple intent detection can be formulated as a multi-label classification task that predicts multiple intents expressed in the input utterance.
And slot filling is a sequence labeling task that maps each $u_i$ into a slot label.

\subsubsection{Model Architecture}
\begin{figure*}[t]
 \centering
 \includegraphics[width = 0.94\textwidth]{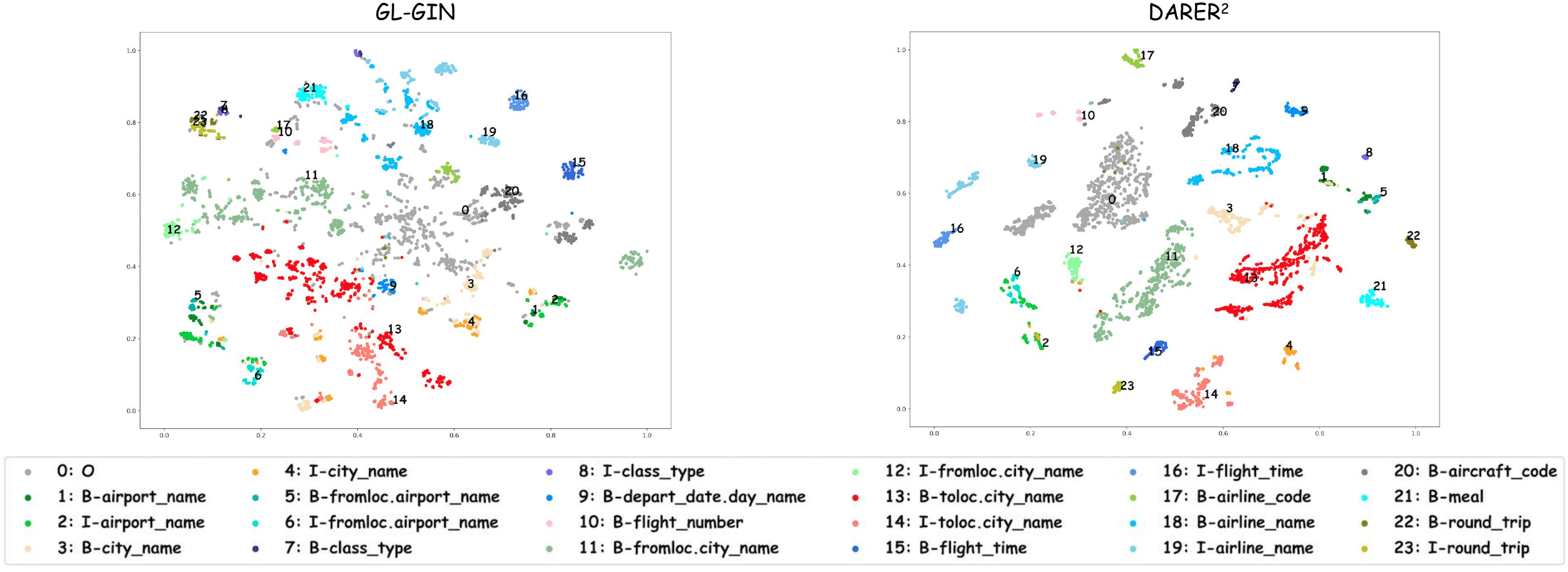}
 \caption{Visualizations of slot hidden states generated by GL-GIN and our DARER$^2$.}
 \label{fig: slot_tsne}
\end{figure*}

We apply our relational temporal graph reasoning to the state-of-the-art model GL-GIN \cite{glgin}, forming DARER and DARER$^2$ in this joint task scenario.
In GL-GIN, the dual-task graph is a semantics-label graph, which is a homogeneous graph including two groups of nodes: predicted intent label nodes and slot semantics nodes.
And vanilla GAT is utilized for graph reasoning.
In our DARER, the dual-task graph is a relational temporal graph, in which there are (1) intra- and cross-task relations among the two tasks' nodes; (2) coarse-grained temporal relations among the slot semantics nodes.
And RGCN is utilized for relational temporal graph reasoning
In our DARER$^2$, the dual-task graph is also a relational temporal graph, including intra- and cross-task relations.
And our proposed DTR-ReTeFormer  is used for relational temporal graph reasoning.
Since DTR-ReTeFormer can achieve the fine-grained temporal modeling, compared with GL-GIN and DARER, DARER$^2$ can capture the dependencies between B- slot labels and their I- slot labels, and this advantage is proven in Fig. \ref{fig: slot_tsne}.

\subsubsection{Datasets and Metrics}
\textbf{Datasets.}Following previous works, the two benchmarks: MixATIS and MixSNIPS  \cite{atis,snips,agif} are used as testbeds for evaluation.
In MixATIS, the split of train/dev/test set is 13162/756/828 (utterances).
In MixSNIPS, the split of train/dev/test set is 39776/2198/2199 (utterances).\\
\textbf{Evaluation Metrics.}
Following previous works, multiple intent detection is evaluated by accuracy (Acc); slot filling is evaluated using F1 score; sentence-level semantic frame parsing is evaluated using overall Acc.
Overall Acc denotes the ratio of utterances for which both intents and slots are predicted correctly.
\subsubsection{Implement Details and Baselines}
Following GL-GIN \cite{glgin}, the word and label embeddings are randomly initialized and trained with the model. 
The dimension of the word/label embedding is 128 on MixATIS and 256 on MixSNIPS.
The dimension of the hidden state is 200.
We adopt Adam \cite{adam} optimizer for model training with the default setting.
For all experiments, we select the best model on the dev set and report its results on the test set.

We compare our model with Attention BiRNN \cite{attbirnn}, Slot-Gated \cite{slot-gated}, Bi-Model \cite{bimodel}, SF-ID \cite{sfid}, Stack-Propagation \cite{qin2019}, Joint Multiple ID-SF \cite{2019-joint}, AGIF \cite{agif} and GL-GIN \cite{glgin}
\subsubsection{Results and Analysis}

\begin{table}[t]
\centering
\fontsize{8}{9}\selectfont
\caption{Results comparison. $^\dag$ denotes our model significantly outperforms baselines with $p<0.01$ under t-test.}
\setlength{\tabcolsep}{0.8mm}{
\begin{tabular}{l|c|c|c|c|c|c}
\toprule
\multirow{2}{*}{Models} & \multicolumn{3}{c|}{MixATIS} & \multicolumn{3}{c}{MixSNIPS} \\ \cline{2-7} 
                        & \begin{tabular}[c]{@{}c@{}}Overall\\(Acc)\end{tabular}  &\begin{tabular}[c]{@{}c@{}}Slot\\(F1)\end{tabular}  &\begin{tabular}[c]{@{}c@{}}Intent\\(Acc)\end{tabular}& \begin{tabular}[c]{@{}c@{}}Overall\\(Acc)\end{tabular}& \begin{tabular}[c]{@{}c@{}} Slot\\(F1)\end{tabular}&\begin{tabular}[c]{@{}c@{}}Intent\\(Acc)\end{tabular}           \\ \midrule
Attention BiRNN \cite{attbirnn} &  39.1 &  86.4      &   74.6      & 59.5        &  89.4     & 95.4 \\
Slot-Gated \cite{slot-gated}    &  35.5 &  87.7      &   63.9      & 55.4        &  87.9     & 94.6 \\
Bi-Model \cite{bimodel}         &  34.4 &  83.9      &   70.3      & 63.4        &  90.7    & 95.6 \\
SF-ID \cite{sfid}               &  34.9 &  87.4      &   66.2      & 59.9        &  90.6     & 95.0 \\
Stack-Propagation \cite{qin2019}&  40.1 &  87.8      &   72.1      & 72.9        &  94.2     & 96.0 \\
Joint Multiple ID-SF \cite{2019-joint}&36.1 &84.6    &   73.4      & 62.9        &  90.6     & 95.1 \\
AGIF \cite{agif}                &  40.8 &  86.7      &   74.4      & 74.2        &  94.2     & 95.1 \\
GL-GIN \cite{glgin}        &  43.0 &  88.2      &   76.3      & 73.7        &  94.0     & 95.7 \\ \midrule
DARER           &  \textbf{44.7}$^\dag$ &\textbf{88.4} &\textbf{76.7}$^\dag$ & \textbf{74.7}$^\dag$  &  \textbf{94.4}$^\dag$ & \textbf{96.5}$^\dag$ \\
DARER$^2$           &  \textbf{49.0}$^\dag$ &\textbf{89.2}$^\dag$ &\textbf{77.3}$^\dag$ & \textbf{76.3}$^\dag$  &  \textbf{94.9}$^\dag$ & \textbf{96.7}$^\dag$ \\
  \bottomrule
\end{tabular}} 
\label{table: jsfid results}
\end{table}

The results comparison is shown in Table \ref{table: jsfid results}.
We can observe that our DARER models significantly outperform the state-of-the-art model GL-GIN on all datasets.
In particular, DARER$^2$ significantly surpasses GL-GIN on MixATIS dataset in terms of Overall Acc, achieving 14\% relative improvement.
The superior performances of our DARER models verify the advantages of our proposed relation temporal graph reasoning.
On one hand, relation temporal graph reasoning can effectively model the intra- and cross-task relation-specific interactions.
On the other hand, it can model the temporal information among the slot semantics nodes.
Especially, the DTR-ReTeFormer in DARER$^2$ can comprehensively model the fine-grained temporal information, capturing the slot dependencies.
To further verify this, we visualize the slot hidden state generated by DARER$^2$ and GL-GIN, as shown in Fig. \ref{fig: slot_tsne}
We can observe that DARER$^2$'s clusters are clearer than GL-GIN's.
Besides, the B- slot clusters of DARER$^2$ and their corresponding I- slot clusters are separated clearly. 
In contrast, some GL-GIN's generated B- slot clusters and their corresponding I- slot clusters even overlap.
The high quality of our DARER$^2$'s generated hidden states can be attributed to three facts.
First, GL-GIN uses the vanilla GAT for information aggregation, leading to a disadvantage: for each slot node, the different information of the intent label nodes and other slot nodes are directly fused to it.
Differently, the DTR-ReTeFormer in DARER$^2$ can achieve relation-specific information aggregation, which can better leverage the beneficial information via discriminating the contributions of the two tasks' nodes.
Second, the GAT in GL-GIN cannot model the temporal information, losing the dependencies among slot nodes.
Since each slot node corresponds to a word in utterance, the group of slot nodes can be regarded as a sequence, where there are temporal dependencies (e.g. I-Singer can only occur behind B-Singer).
And our DARER models can achieve the relational temporal modeling, then capture the beneficial slot dependencies.

\section{Conclusion and Future Work}\label{sec: conclusion}
In this paper, we present a new framework, which for the first time achieves relational temporal graph reasoning and integrates prediction-level interactions to leverage estimated label distribution as explicit and important clues other than implicit semantics.
We design the SATG and DRTG to facilitate relational temporal graph reasoning of dialog understanding and dual-task reasoning.
To achieve our framework, we first propose a novel model named DARER to model the relational interactions between temporal information, label information, and semantics to let two tasks gradually promote each other, which is further forced by the proposed logic-heuristic training objective.
Then we propose DARER$^2$, which further enhances relational temporal graph reasoning by adopting our proposed SAT-ReTeFormer and DTR-ReTeFormer.
Experimental results demonstrate the superiority of our DARER models, which surpasses previous models by a large margin in different dual-task dialog language understanding scenarios.

Our work brings two insights for dialog understanding and multi-task reasoning in dialog systems: (1) exploiting the relational temporal information of the dialog for graph reasoning; (2) leveraging estimated label distributions to capture explicit correlations between the multiple tasks.
In the future, we will apply our method to other multi-task learning scenarios in dialog systems.

\ifCLASSOPTIONcompsoc
  \section*{Acknowledgments}
\else
  \section*{Acknowledgment}
\fi

This work was supported by Australian Research Council Grant DP200101328.
Bowen Xing and Ivor W. Tsang was also supported by A$^*$STAR Centre for Frontier AI Research.

\ifCLASSOPTIONcaptionsoff
  \newpage
\fi

\normalem
\bibliographystyle{IEEEtran}
\bibliography{anthology.bib}
%
%

\begin{IEEEbiography}[{\includegraphics[width=1in,height=1.25in,clip,keepaspectratio]{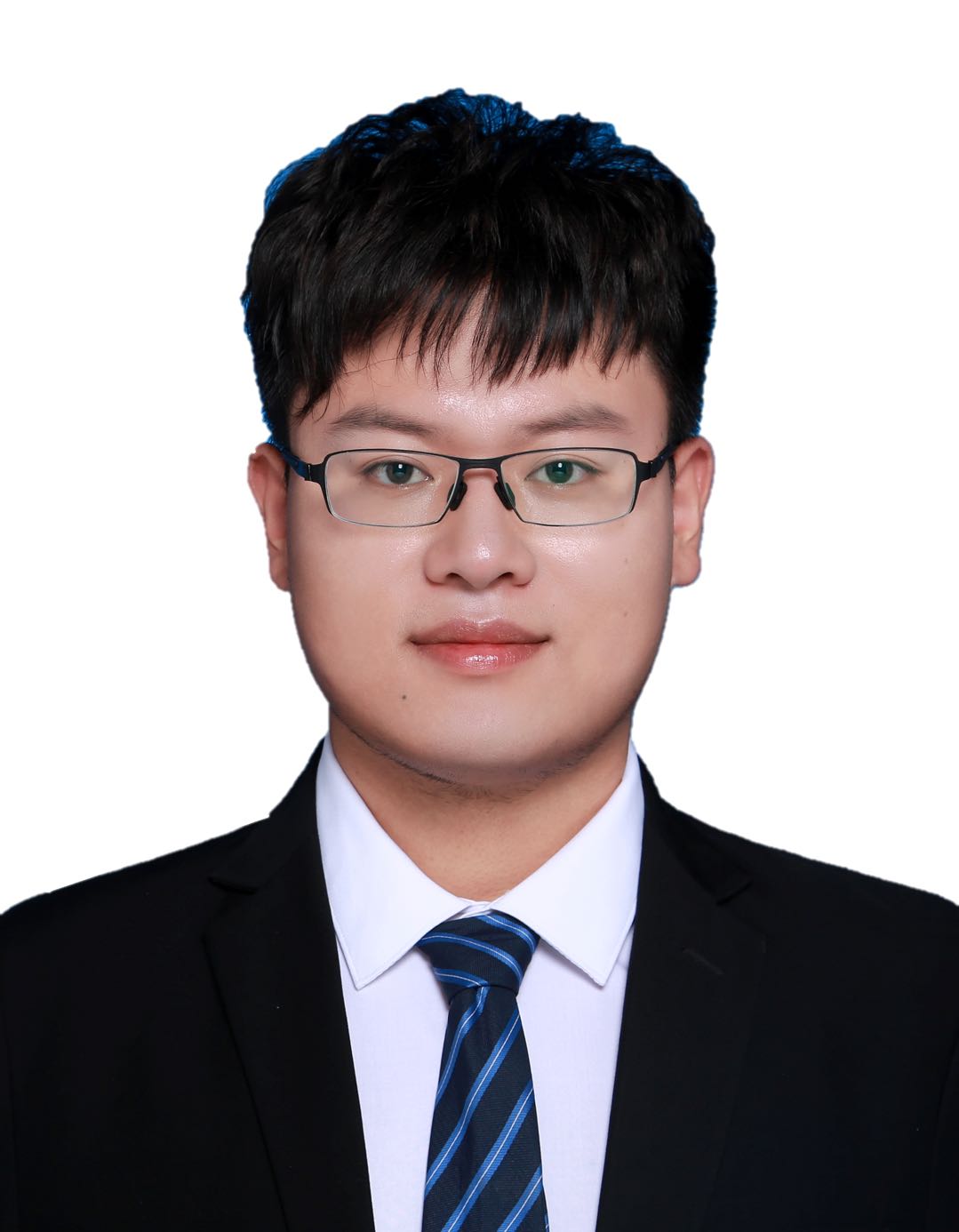}}]{Bowen Xing}
received his B.E. degree and Master degree from Beijing Institute of Technology, Beijing, China, in 2017 and 2020, respectively. He is currently a third-year Ph.D student at or of the Australian Artificial Intelligence Institute (AAII), University of Technology Sydney (UTS).
His research focuses on graph neural networks, multi-task learning, sentiment analysis, and dialog system.
\end{IEEEbiography}

\begin{IEEEbiography}[{\includegraphics[width=1in,height=1.25in,clip,keepaspectratio]{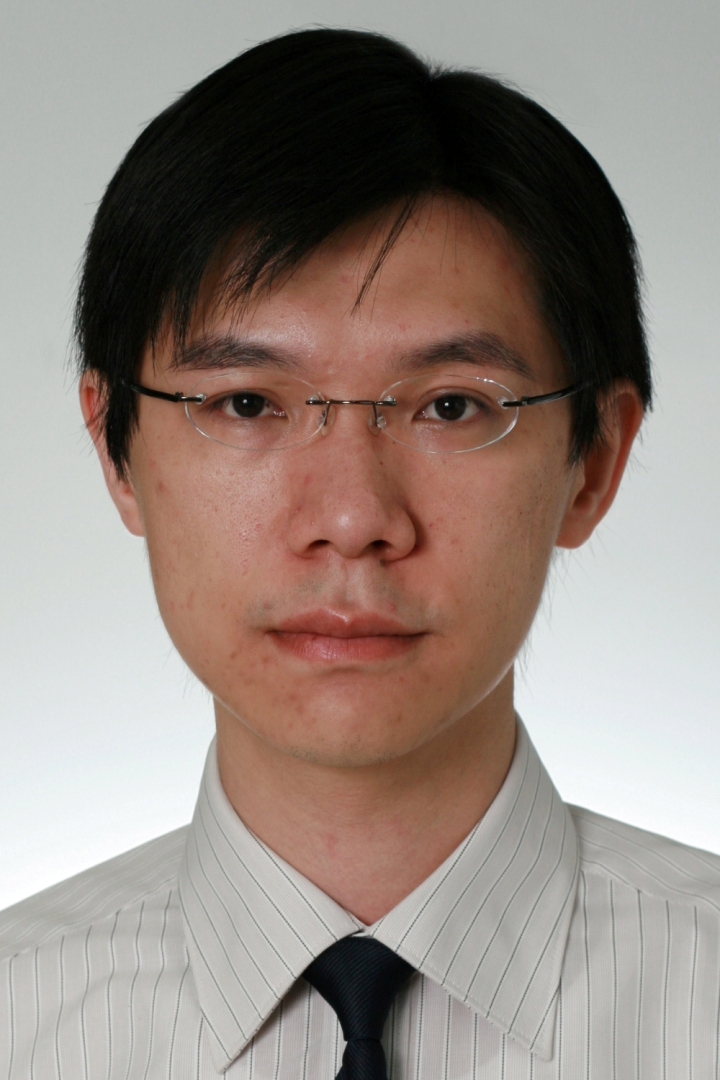}}]{Ivor W. Tsang} is an IEEE Fellow and the Director of A*STAR Centre for Frontier AI Research (CFAR). Previously, he was a Professor of Artificial Intelligence, at University of Technology Sydney (UTS), and Research Director of the Australian Artificial Intelligence Institute (AAII).
His research focuses on transfer learning, deep generative models, learning with weakly supervision, big data analytics for data with extremely high dimensions in features, samples and labels. His work is recognised internationally for its outstanding contributions to those fields.
In 2013, Prof Tsang received his ARC Future Fellowship for his outstanding research on big data analytics and large-scale machine learning. 
In 2019, his JMLR paper ``Towards ultrahigh dimensional feature selection for big data'' received the International Consortium of Chinese Mathematicians Best Paper Award. In 2020, he was recognized as the AI 2000 AAAI/IJCAI Most Influential Scholar in Australia for his outstanding contributions to the field, between 2009 and 2019. His research on transfer learning was awarded the Best Student Paper Award at CVPR 2010 and the 2014 IEEE TMM Prize Paper Award. In addition, he received the IEEE TNN Outstanding 2004 Paper Award in 2007 for his innovative work on solving the inverse problem of non-linear representations. Recently, Prof Tsang was conferred the IEEE Fellow for his outstanding contributions to large-scale machine learning and transfer learning.
Prof Tsang serves as the Editorial Board for the JMLR, MLJ, JAIR, IEEE TPAMI, IEEE TAI, IEEE TBD, and IEEE TETCI. He serves as a Senior Area Chair/Area Chair for NeurIPS, ICML, AAAI and IJCAI, and the steering committee of ACML.
\end{IEEEbiography}

\end{document}